\documentclass[10pt,conference]{IEEEtran}

\usepackage{booktabs} 
\usepackage[flushleft]{threeparttable}
\usepackage{xcolor} 
\usepackage[utf8]{inputenc}
\usepackage{float}
\usepackage{amssymb}
\usepackage{amsmath}
\usepackage{ifthen}
\usepackage{multirow} 
\usepackage{caption}
\usepackage{listings}
\usepackage[bookmarks=false]{hyperref}
\usepackage{url,moreverb,xspace}
\usepackage{tcolorbox}
\usepackage{xspace}
\usepackage{cite}

\usepackage{enumitem}
\usepackage{array,graphicx}
\usepackage{soul}
\usepackage{balance}
\usepackage{pifont}
\usepackage{etoolbox,siunitx}
\usepackage{nicefrac}
\usepackage{mathtools}

\usepackage{colortbl,xcolor,pifont}
\newcommand*\colourcheck[1]{%
  \expandafter\newcommand\csname #1check\endcsname{\textcolor{#1}{\ding{52}}}%
}
\newcommand*\colourcross[1]{%
  \expandafter\newcommand\csname #1cross\endcsname{\textcolor{#1}{\ding{56}}}%
}
\usepackage[vlined, boxruled, linesnumbered] {algorithm2e}
\SetKw{KwBy}{by}
\SetKw{KwBreak}{break}
\SetKw{KwReturn}{return}
\def\HiLi{\leavevmode\rlap{\hbox to \hsize{\color{gray!35}\leaders\hrule height .8\baselineskip depth .5ex\hfill}}}

\definecolor{lightgray}{gray}{0.85}
\definecolor{bananayellow}{rgb}{1.0, 0.88, 0.21}

\usepackage{graphicx}
\usepackage{subcaption}
\usepackage{seqsplit}

\newcommand{\selforacle}{SelfOracle\xspace} %
\newcommand{\ty}{ThirdEye\xspace} %
\newcommand{\davetwo}{\mbox{DAVE-2}\xspace} %
\newcommand{\numsim}{70\xspace} %
\newcommand{\mcdropout}{MC-Dropout\xspace} %

\SetKwComment{Comment}{$\triangleright$\ }{}
\SetCommentSty{itshape}
\newboolean{showcomments}
\setboolean{showcomments}{true}
\ifthenelse{\boolean{showcomments}}
{\newcommand{\nb}[2] {
  \fcolorbox{black}{gray!20}{\bfseries\sffamily\scriptsize#1:}
  {\sf\small$\blacktriangleright$\textit{#2}$\blacktriangleleft$}
}
}
{\newcommand{\nb}[2]{}
}

\newcounter{fcounter}
\newcommand{\curl}[1]{\footnote{\url{#1}}}

\makeatletter
\newcommand{\thickhline}{%
    \noalign {\ifnum 0=`}\fi \hrule height 1pt
    \futurelet \reserved@a \@xhline
}
\makeatother

\begin{document}

\pagenumbering{arabic} 
\pagestyle{plain}

\title{Predicting Safety Misbehaviours in Autonomous Driving Systems using Uncertainty Quantification}

\author{\IEEEauthorblockN{Ruben Grewal}
\IEEEauthorblockA{
\textit{Technical University of Munich}\\
Munich, Germany \\
ruben.grewal@tum.de}
\and
\IEEEauthorblockN{Paolo Tonella}
\IEEEauthorblockA{
\textit{Software Institute - USI}\\
Lugano, Switzerland \\
paolo.tonella@usi.ch}
\and
\IEEEauthorblockN{Andrea Stocco}
\IEEEauthorblockA{
\textit{Technical University of Munich, fortiss GmbH}\\
Munich, Germany \\
andrea.stocco@tum.de, stocco@fortiss.org}
}


\IEEEoverridecommandlockouts
\IEEEpubid{\makebox[\columnwidth]{978-1-5386-5541-2/ 20/\$31.00~\copyright2024 IEEE \hfill} \hspace{\columnsep}\makebox[\columnwidth]{ }}

\maketitle

\IEEEpubidadjcol

\begin{abstract}
The automated real-time recognition of unexpected situations plays a crucial role in the safety of autonomous vehicles, especially in unsupported and unpredictable scenarios. This paper evaluates different Bayesian uncertainty quantification methods from the deep learning domain for the anticipatory testing of safety-critical misbehaviours during system-level simulation-based testing. Specifically, we compute uncertainty scores as the vehicle executes, following the intuition that high uncertainty scores are indicative of unsupported runtime conditions that can be used to distinguish safe from failure-inducing driving behaviors. 
In our study, we conducted an evaluation of the effectiveness and computational overhead associated with two Bayesian uncertainty quantification methods, namely MC-Dropout and Deep Ensembles, for misbehaviour avoidance. Overall, for three benchmarks from the Udacity simulator comprising both out-of-distribution and unsafe conditions introduced via mutation testing, both methods successfully detected a high number of out-of-bounds episodes providing early warnings several seconds in advance, outperforming two state-of-the-art misbehaviour prediction methods based on autoencoders and attention maps in terms of effectiveness and efficiency. 
Notably, Deep Ensembles detected most misbehaviours without any false alarms and did so even when employing a relatively small number of models, making them computationally feasible for real-time detection. Our findings suggest that incorporating uncertainty quantification methods is a viable approach for building fail-safe mechanisms in deep neural network-based autonomous vehicles.
\end{abstract}

\begin{IEEEkeywords}
autonomous vehicles testing, uncertainty quantification, self-driving cars, failure prediction.
\end{IEEEkeywords}

%
%




\section{Introduction}\label{sec:introduction}

Autonomous driving systems (ADS) are vehicles equipped with sensors, cameras, radar, and artificial intelligence, used to let them travel between destinations without human intervention. For a vehicle to be qualified as fully autonomous, it must possess the capability to autonomously navigate to a predefined destination on roads that have not been specifically adapted for its use~\cite{yurtsever2020survey}.
The U.S. Department of Transportation, National Highway Traffic Safety Administration (NHTSA), has defined five standardized levels of autonomy, from driver assistance (with the driver being responsible for safe driving) to full automation (where no human driver is required to operate the vehicle).
Several companies, such as Audi, BMW, Ford, Google, General Motors, Tesla, Volkswagen, and Volvo, are actively engaged in the development and testing of autonomous vehicles. In recent years, we witnessed advancements such as people hailing self-driving taxis or fleets of fully automated cars with no accompanying safety drivers.
Deep neural networks (DNNs) are the driving force behind self-driving car systems. To create autonomous vehicles, developers rely on extensive datasets harnessed in the field to train large DNNs. This data includes images captured by cameras on actual vehicles and other sensors, enabling the DNNs to learn to identify road elements, traffic lights, pedestrians, and other elements within diverse driving environments~\cite{grigorescu2020survey}.

Safety assessment of ADS is a hard endeavor and extensive testing is required before deployment on public roads. To validate the safety of ADS, companies adopt a multi-pillar approach that encompasses simulation-based testing, test track, and real-world testing~\cite{waymos-secret-testing,Cerf:2018:CSC:3181977.3177753}. 
Researchers have focused primarily on the first pillar, proposing automated testing techniques that try to expose failing conditions and corner cases~\cite{Gambi:2019:ATS:3293882.3330566,Abdessalem-ICSE18,Abdessalem-ASE18-1,Abdessalem-ASE18-2,2020-Riccio-FSE}.
However, despite these efforts, public acceptance of autonomous driving software in the real world would consider the capabilities of the ADS to operate safely in partially unknown and uncertain environments, therefore exhibiting a high level of robustness also for sensor inaccuracies and environmental uncertainties~\cite{NIPS2015_86df7dcf}. 

DNNs are known for their tendency to produce unexpectedly incorrect yet overly confident predictions, particularly in complex environments like autonomous driving. This poses significant safety concerns for ADS, which should possess situational awareness capabilities to discern challenging scenarios, such as adverse weather conditions, which are likely to induce errors and then prompt timely warnings to the driver or trigger fail-safe mechanisms~\cite{he2023survey,Weiss2021FailSafe}.

Previous research has introduced techniques to build safety in-service monitoring~\cite{Henriksson,Kim:2019:GDL:3339505.3339634,9402003,deeproad,dissector,2020-Stocco-ICSE,knoll-monitoring}. 
Frameworks such as SelfOracle~\cite{2020-Stocco-ICSE}, DeepRoad~\cite{deeproad}, DeepGuard~\cite{DeepGuard} require a data-box access~\cite{2020-Riccio-EMSE} and they are capable of analyzing real-world driving data and assess whether the ADS is safe. However, these approaches work in a \textit{black-box} manner (i.e., they analyze the input/output data and identify anomalous instances, without considering the internal processing by the DNN model), which makes them less sensitive to bugs at the model level~\cite{2020-Humbatova-ICSE} and prone to false positives/negatives, given their external perspective on the system being tested. 
A recent white-box solution uses attention maps as a proxy of the DNN uncertainty to enhance the accuracy of failure prediction~\cite{2022-Stocco-ASE}, but it comes with higher costs and therefore is less suitable for resource-constrained environments.

This paper investigates the problem of building a white-box ADS failure predictor rooted in the uncertainty quantification (UQ) methods available in the deep learning domain. 
Uncertainty quantification consists of approaches that compute the confidence, or lack thereof, of deep learning models in response to certain inputs~\cite{he2023survey}. UQ is widely used for the analysis, testing, comprehension, and debugging of DNNs~\cite{he2023survey,Weiss2021FailSafe}. 
In this work, we evaluate two UQ methods for failure prediction to keep the reliability of the ADS within safety bounds. Our approach leverages
uncertainty scores as a transparent confidence estimator for the system.
Online monitoring is performed during ADS driving; the uncertainty scores synthesized from the internals of the DNN under test are used to automatically identify conditions in which the system is not confident. In this paper, we show that uncertainty scores represent important clues about the reliability of the ADS and can be used as failure predictors. Our technique works unsupervisedly as failure prediction is performed by establishing a threshold over the uncertainty scores during nominal operating conditions. Hence, anomalous driving conditions are detected when the uncertainty scores increase above such threshold within a specific detection window preceding the failure.

We have evaluated the effectiveness of uncertainty quantification methods on the Udacity simulator for self-driving cars~\cite{udacity-simulator}, using ADS available from the literature and a diverse set of failures induced by adverse operational scenes and mutation testing-simulated malfunctions. 
More specifically, we evaluated two uncertainty quantification methods (i.e., Monte Carlo Dropout and Deep Ensembles) and their effectiveness when varying their hyperparameters (e.g., number of models or samples used for uncertainty estimation) at different confidence levels. 
In our experiments using an existing dataset of +\numsim simulations accounting for more than 250 failures~\cite{2022-Stocco-ASE}, UQ methods demonstrated remarkable predictive capabilities, forecasting most failures several seconds in advance, a 6-15\% increase in failures detected compared to \selforacle~\cite{2020-Stocco-ICSE} and \ty~\cite{2022-Stocco-ASE}, two state-of-the-art strategies from the literature based on autoencoders and attention maps. 
Notably, our most successful UQ method strikes a superior balance between identifying misbehaviors and minimizing false alarms (94\% $F_{3}$ score) for a relatively constrained configuration, ensuring computational feasibility for real-time detection.

Our paper makes the following contributions:

\begin{description}[noitemsep]
\item [Technique.] A monitoring technique for ADS failure prediction based on uncertainty quantification methods. Our approach is publicly available as a tool~\cite{replication-package}. 
\item [Evaluation.] An empirical study showing that the uncertainty scores are a promising white-box confidence metric for failure prediction, outperforming the black-box approach of \selforacle~\cite{2020-Stocco-ICSE} and the XAI-based approach by \ty~\cite{2022-Stocco-ASE}. Our study also discusses the performance of our methods for real-time prediction.
\end{description}

\section{Background}\label{sec:background}

\subsection{Lane-keeping ADS}

ADS rely on sensor data, cameras, and GPS to perceive their surroundings and use different processing methods to enable predictive decisions regarding vehicle controls~\cite{yurtsever2020survey}. 

From an architectural point of view, ADS can be mainly divided into two categories: end-to-end ADS driving models and multi-module ADS. The former ones are based on advanced DNNs that are trained on massive datasets of driving scenes. 
The latter ones are organized into four modules: perception, prediction, planning, and control~\cite{yurtsever2020survey}. The perception module receives as input various sources of sensor data, such as images of the front camera, and proximity sensor, to detect objects in the neighborhood of the vehicle. The prediction module predicts the trajectories of these objects, which are used by the planning module to decide a safe route. The control module translates the route into actual vehicle commands, e.g., a sequence of steering angles. As of now, the two approaches coexist~\cite{yurtsever2020survey} and it is not clear if an approach will prevail. 

In this paper, we consider testing end-to-end ADS, while we leave the investigation of multi-module ADS for future work. Particularly, we focus on ADS that implement the ``behavioral cloning'' task through imitation learning. In this task, the vehicle learns the function of lane-keeping in an end-to-end manner, from human-labeled driving samples in which actuators' values reflect the driving decisions of an expert human driver operating a real physical vehicle, or a simulated vehicle within a driving simulator~\cite{udacity-simulator}. Once trained, models like NVIDIA's \davetwo~\cite{nvidia-dave2} are capable of predicting the vehicle's controls (i.e., steer, brake, acceleration).

The ability to keep the vehicle within a lane is a fundamental component of the safe deployment of DNN-based ADS. Notably, the NHTSA has reported that off-road failures are not only frequent but also come at a significant cost, exceeding 15 billion USD~\cite{precrashreport}.

\subsection{Failure Conditions for Lane-keeping ADS}\label{sec:failures}

In the context of NHTSA Level 4 (High Automation), a system monitor plays a critical role in identifying emerging functional insufficiencies. Its primary objective is to maintain a high level of functional quality, even in extreme situations~\cite{dissector,deeproad,Henriksson}. When the monitor deems the current condition as unsafe, the ADS should be designed to disengage, requesting human intervention to take control of the vehicle, or activating alternative fail-safe mechanisms~\cite{Weiss2021FailSafe}.

\begin{figure}[t]
\centering
\includegraphics[trim={0cm 8cm 0cm 8cm}, clip, width=0.3\columnwidth]{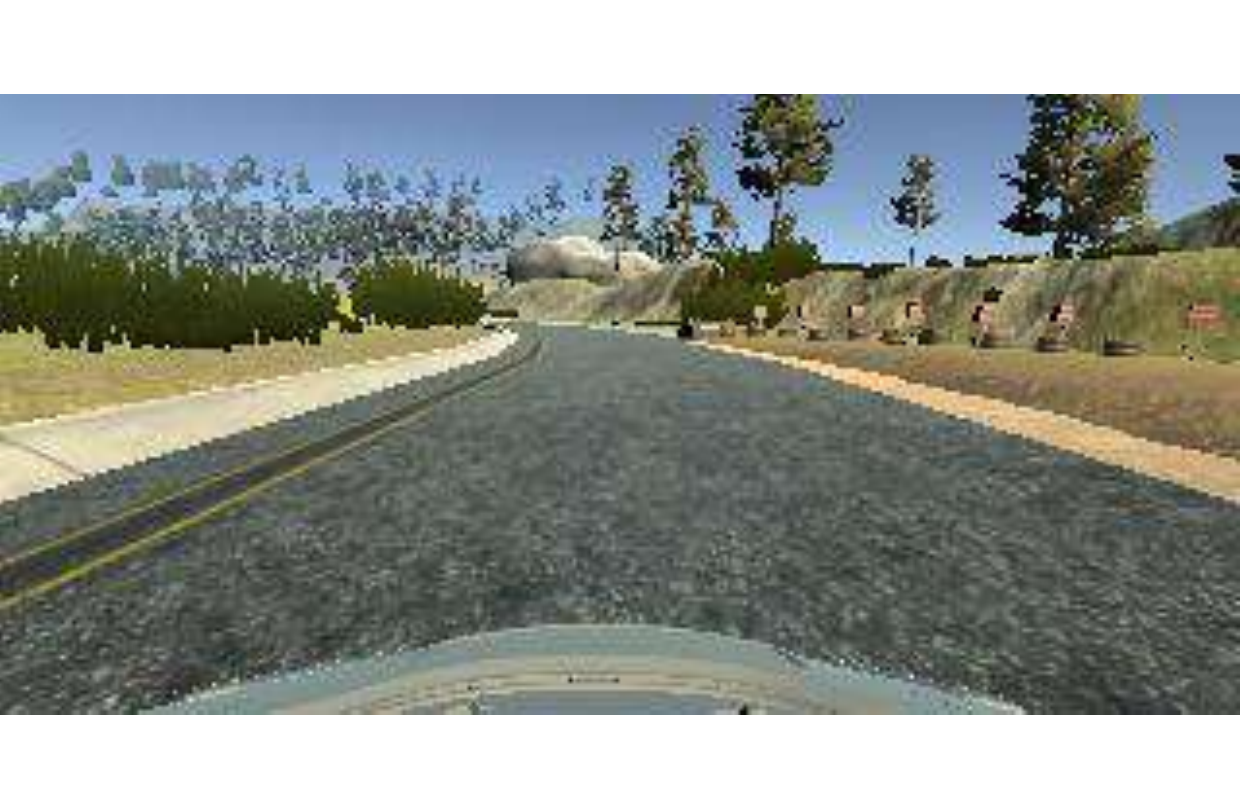}
\includegraphics[trim={0cm 8cm 0cm 8cm}, clip, width=0.3\columnwidth]{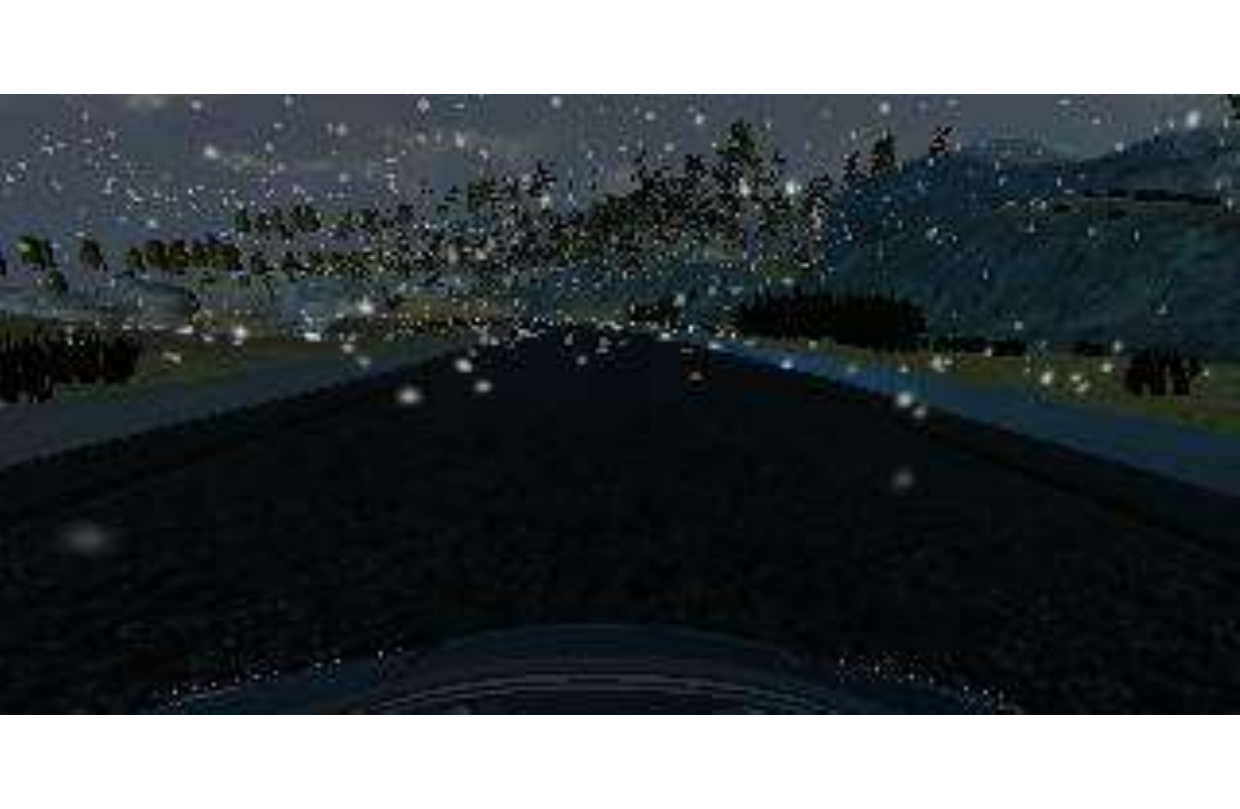}
\includegraphics[trim={0cm 8cm 0cm 8cm}, clip, width=0.3\columnwidth]{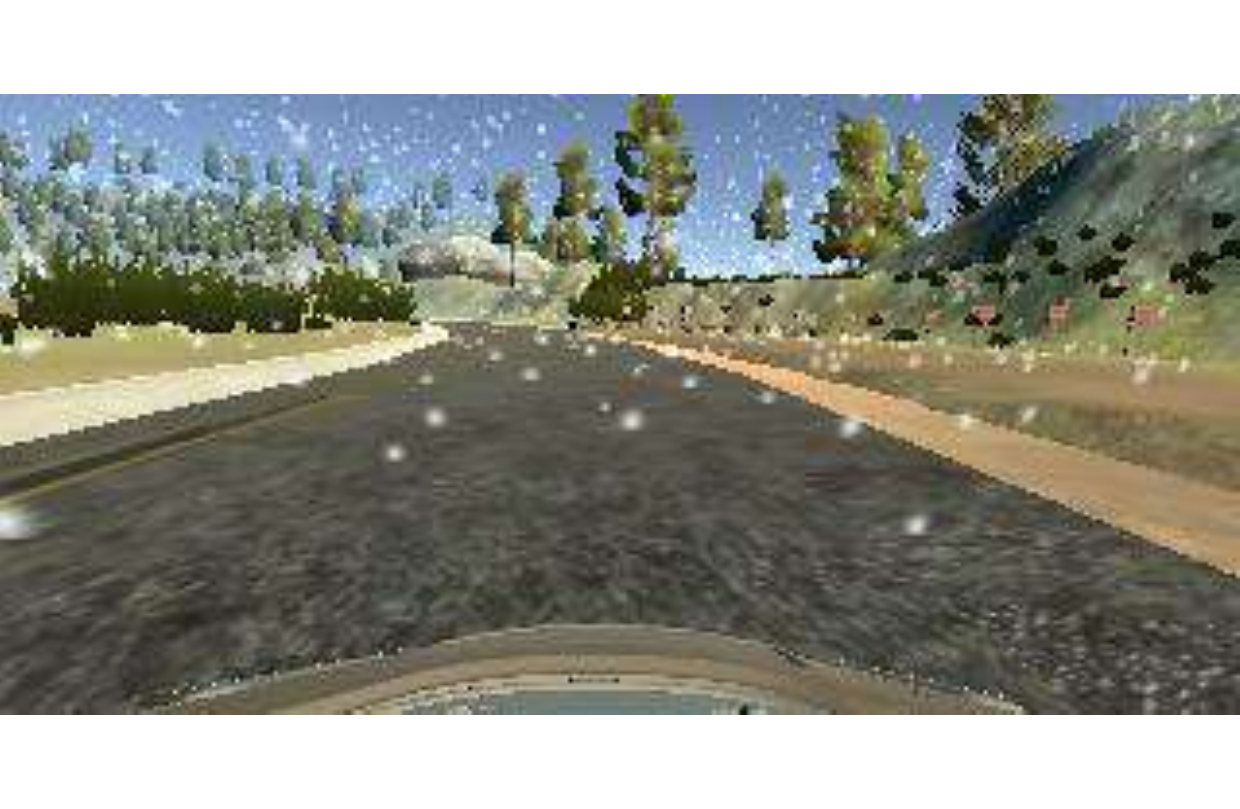}
\caption{Examples of operational conditions~\cite{2022-Stocco-ASE}. \ul{Left}: nominal (sunny). \ul{Center}: OOD (night+snow). \ul{Right}: OOD (snow).}
\label{fig:conditions}
\end{figure}

Among the underlying causes of ADS failures, such as instances of off-road driving, SOTIF ~\cite{sotif} highlights the role of both external unknown and internal uncertain conditions~\cite{sotif}. External unknown conditions encompass ``abnormal'' inputs that represent rare, unexpected, and potentially unsupported environmental events. These conditions typically involve scenarios where the ADS was not trained due to the absence of prior knowledge (i.e., epistemic uncertainty), such as specific road types or particular weather and lighting conditions. The DNNs utilized within ADS may not be resilient to such significant changes in data distribution and they are said to be out-of-distribution (OOD, see \autoref{fig:conditions}), potentially resulting in system-level failures such as the ADS driving off-road.
Conversely, internal uncertain conditions pertain to misbehaviors within the decision component of the ADS. These misbehaviors are often attributed to inherent bugs in the DNN model, which may be introduced during its development phase. Common instances of such bugs include inadequate training data and sub-optimal choices regarding the model architecture or training hyperparameters~\cite{2020-Humbatova-ICSE}. In the rest of the paper, we shall use the terms failures/misbehaviours interchangeably.

\subsection{Existing Unsupervised Failure Predictions Methods}\label{sec:existing-unsup-failure-prediction}

Researchers have proposed ADS failure prediction models that can be trained with no supervision (i.e., no knowledge of the anomalies). Certain propositions are based on a data-box access~\cite{2020-Riccio-EMSE}\footnote{In the original papers, these solutions are described as black-box methods, despite their reliance on access to the training set of the ADS. Therefore, it would be more accurate to consider them as data-box techniques. However, for the sake of simplicity, this paper employs the term black-box to refer to the existing data-box techniques that are applied in a black-box manner.} to the main system~\cite{2020-Stocco-ICSE,DeepGuard,knoll-monitoring,Henriksson}, whereas other solutions require internal information of the systems and therefore are considered white-box~\cite{Michelmore,MichelmoreWLCGK20,2022-Stocco-ASE}. 

In this work, we chose two representative propositions from both domains, namely \selforacle~\cite{2020-Stocco-ICSE} and \ty~\cite{2022-Stocco-ASE}. 
We selected these approaches as baselines because they represent two competitive approaches, one black-box, and one white-box, that are designed for the task of failure prediction of ADS and use an unsupervised failure predictor to analyze inputs and assign a suspiciousness score to them, which should be low (below a threshold) if the inputs are supported, or high (above a threshold) otherwise. 

These approaches were developed, integrated, and experimented on the Udacity simulator~\cite{udacity-simulator}. 
In this paper, we evaluate our failure predictors in the same experimental setting as previous work to mitigate the threats to the internal validity that are possible when experimenting with tools in a simulation environment different from the one in which they were originally implemented. In the following of this section, we provide further details on the two baseline approaches.

\selforacle~\cite{2020-Stocco-ICSE} is a black-box technique that estimates the system confidence by analyzing the front-facing camera images used by the ADS. \selforacle uses an autoencoder to reconstruct driving images and the reconstruction loss as a measure of confidence. The autoencoder is trained to minimize the distance between the original data and its low-dimensional reconstruction with metrics such as the Mean Squared Error (MSE). A low MSE indicates that the input has characteristics similar to those of the training set, whereas a high MSE indicates potentially an unsupported sample. 
While effective, the main criticism of \selforacle is that it is not informed by the internal functioning of the DNNs responsible for controlling the ADS, as its only connection with such DNNs is the common training set (i.e., the same inputs are used to train DNNs and autoencoder, which makes these or similar inputs relatively familiar and easy to handle/reconstruct for both DNNs/autoencoder). 

To address this, \ty~\cite{2022-Stocco-ASE} was proposed as a white-box alternative based on the attention maps produced by explainable artificial intelligence techniques (XAI). \ty synthesizes suspiciousness scores using different strategies (i.e., pixel-level average, or autoencoder-based reconstruction loss). While proved promising, such confidence scores are only a proxy of the true uncertainty. Second, computing heatmaps at runtime requires a non-negligible computational overhead, which makes their application as a runtime monitoring prediction system a careful, if at all possible, choice.

In this paper, we aim to ground the benefits of UQ for misbehaviour prediction and compare them with such existing approaches. While uncertainty quantifiers are expected to be informative as they are based on full access to the DNN's internals, they are also known to be computationally expensive. To the best of our knowledge, no empirical comparison has been conducted concerning their effectiveness and efficiency, which represent the core objectives of this work.

\section{Deep Neural Networks Uncertainty Quantification Methods}\label{sec:uq-methods}

Uncertainty quantification has gained an increasingly pivotal role in ensuring the reliability and robustness of DNNs, especially those tasked with making critical decisions. 
Uncertainty can be classified into two main types: aleatoric uncertainty and epistemic uncertainty~\cite{hullermeier2021aleatoric}. Aleatoric uncertainty arises because of the random nature of the system under study, while epistemic uncertainty stems from the lack of knowledge of the system. Aleatoric uncertainty cannot be reduced but can be identified and quantified. Conversely, epistemic uncertainty can be reduced through methods such as sensitivity analysis~\cite{bjarnadottir2019climate}, re-training, and fine-tuning. The total predictive uncertainty can be regarded as the sum of aleatoric and epistemic uncertainty~\cite{he2023survey}.

In the following, we summarize two popular UQ methods proposed in the literature, namely Monte Carlo dropout and Deep Ensembles, and their significance in supervising regression DNNs, such as the ones employed for ADS ~\cite{Weiss2021FailSafe,Weiss2021UncertaintyWizard}

\subsection{Monte Carlo Dropout}\label{sec:mc-dropout}

\begin{figure}[t]
\centering
\includegraphics[trim={2cm 1cm 3cm 1cm}, clip, width=0.4\textwidth]{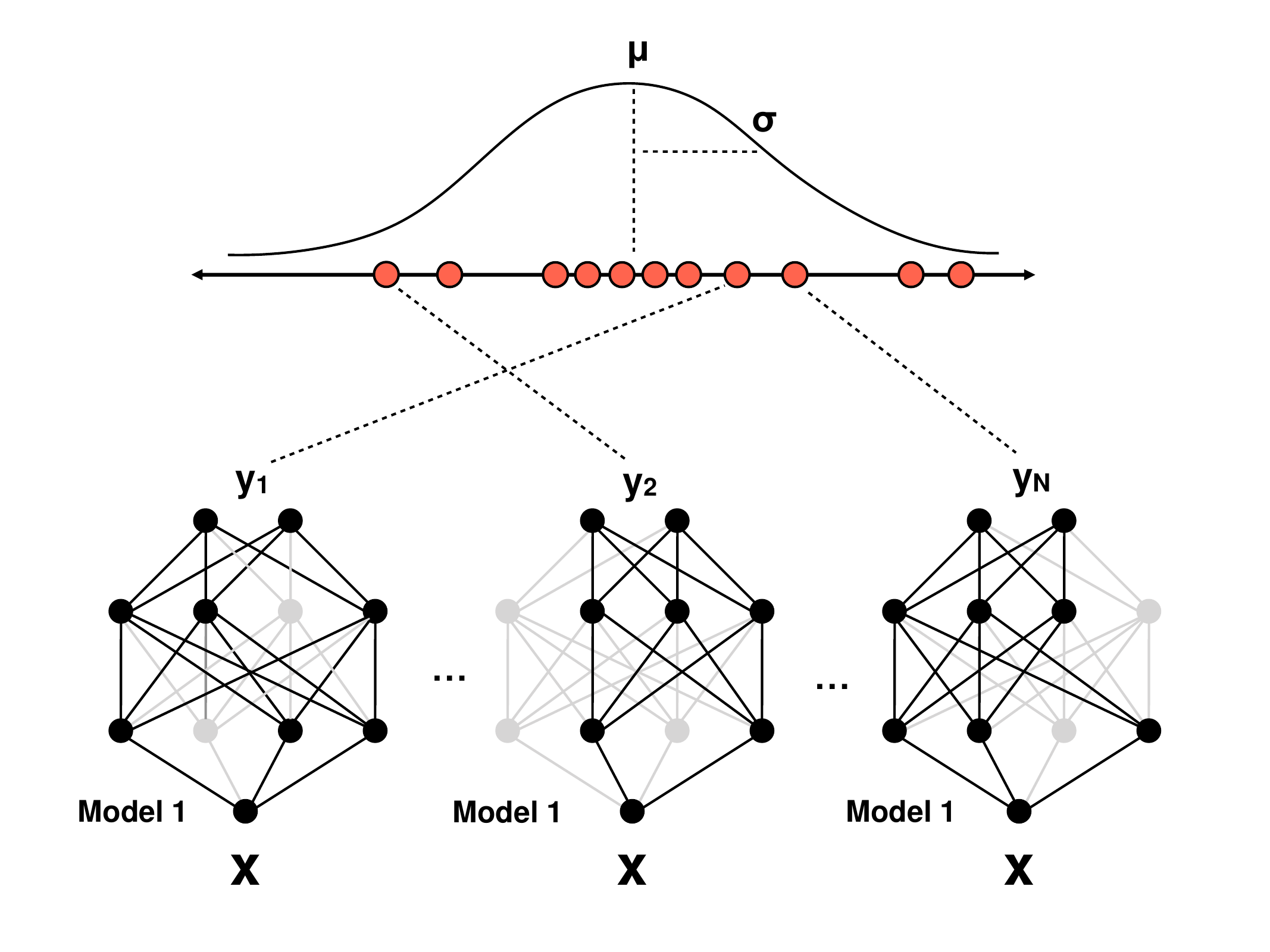}
\caption{Distribution approximated through \mcdropout.}
\label{fig:mc-dropout}
\end{figure}

The first considered UQ method is Monte Carlo Dropout (MC-Dropout or MDC for short)~\cite{10.5555/3045390.3045502}. 
In DNNs, dropout layers are used at training time as a regularization method to avoid overfitting. At testing time they are usually disabled for efficiency reasons, and the final DNN prediction would be deterministic.
However, uncertainty-aware DNNs based on MCD can be enabled based on the principle of Markov Chain Monte Carlo. When estimating predictive uncertainty with \mcdropout, the dropout layers of the DNN are enabled also at inference time. Hence, predictions are no longer deterministic, being dependent on which nodes/links are randomly chosen by the network (see \autoref{fig:mc-dropout}). Therefore, given the same test data point ($X$ in the figure), the model will predict slightly different values every time the point is processed by the DNN, by ``dropping'' a selection of neurons across layers, except for the output layer. 

This method can be regarded as an approximate Bayesian Neural Network (BNN) approach to uncertainty modeling. The Bayesian approach defines the model's likelihood, where Gaussian likelihood is often assumed for regression, with $\omega$ being the model parameters, $x$ the input and $y$ the output~\cite{abdar2021review}:

\begin{equation*}
\centering
    p(y|x, \omega) = \mathcal{N}(avg(f_{\omega}(x)), var(f_{\omega}(x)))
\end{equation*}

MCD is used to generate samples interpreted as a probability distribution through Bayesian interpretation~\cite{10.5555/3045390.3045502                       }: the value predicted by the DNN will be the mean ($avg$, or $\mu$ in \autoref{fig:mc-dropout}) of such probability distribution. Moreover, by collecting multiple predictions for input, each with a different realization of weights due to dropout layers, it is possible to account for model uncertainty as the variance ($var$, or $\sigma$ in \autoref{fig:mc-dropout}) of the observed probability distribution. 

The rationale for using \mcdropout is that supported inputs are expected to be characterized by low DNN uncertainties, whereas unsupported inputs are expected to increase it~\cite{2021-Stocco-JSEP}.

While being simple to implement, \mcdropout is an intrusive approach, as it requires access to the existing DNN architectures, for which dropout layers need to be enabled also at testing time, or added if not already present~\cite{he2023survey}.

Two hyperparameters influence the behaviour of MCD: (i)~the number of stochastic forward passes and (ii)~the dropout rate. While empirical guidelines exist~\cite{10.5555/3045390.3045502}, in this paper we aim to assess the effectiveness of MCD as a failure predictor for ADS testing under a large combination of these parameters. 

\subsection{Deep Ensembles}\label{sec:deep-ensembles}

\begin{figure}[t]
\centering
\includegraphics[trim={2cm 1cm 3cm 1cm}, clip, width=0.4\textwidth]{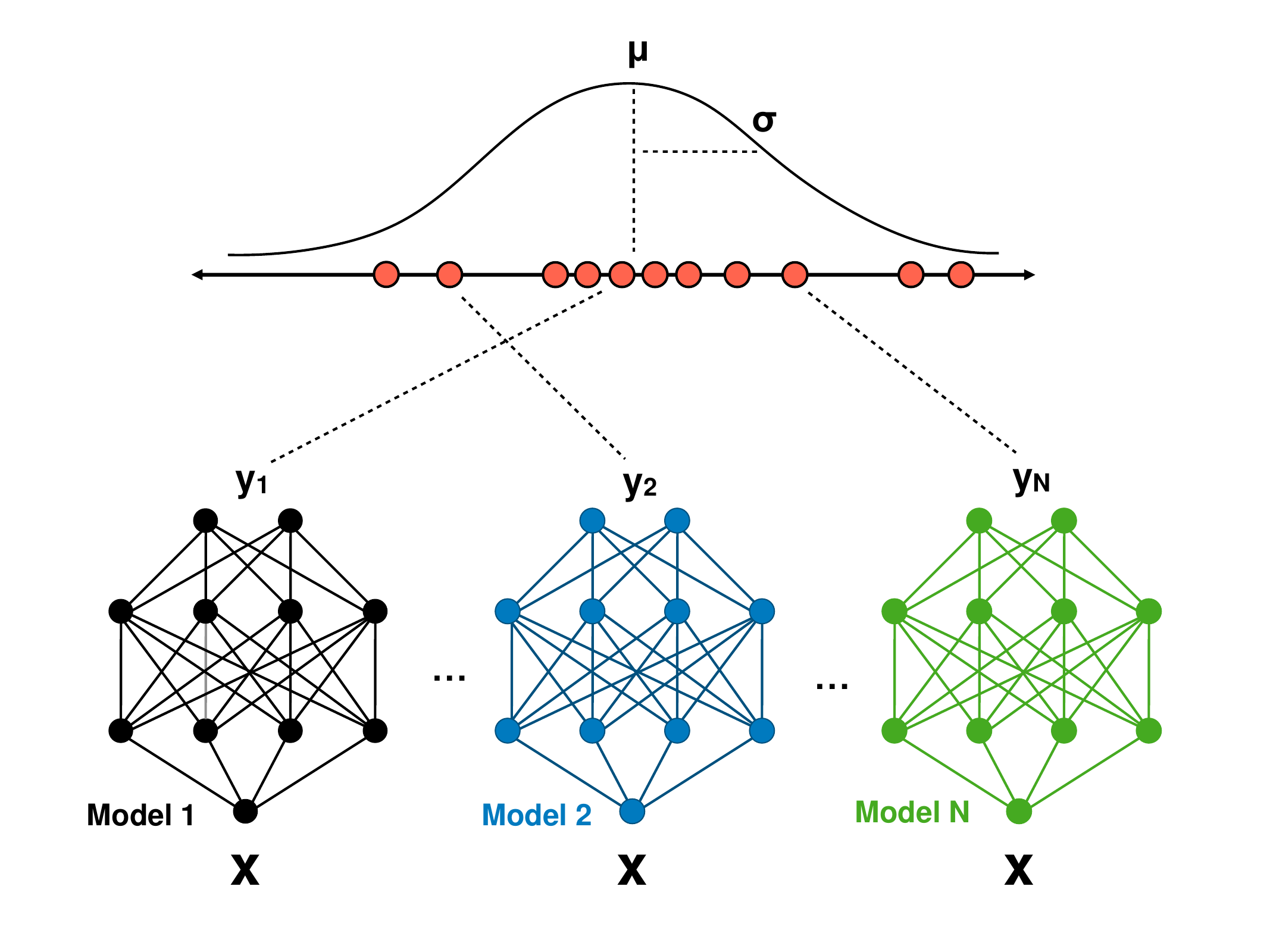}
\caption{Distribution approximated through Deep Ensembles.}
\label{fig:de}
\end{figure}

The second considered UQ method involves another Bayesian method called Deep Ensembles~\cite{lakshminarayanan2017simple} (DE). 
DE requires training multiple instances of the same model architecture on the same dataset while varying other factors to introduce randomicity. The ensemble predictions constitute an output distribution in which the variance of the ensemble characterizes the uncertainty (i.e., a larger variance implies larger uncertainty). 
Among the strategies to build DE, we recall bootstrapping, using different DNN
architectures in terms of a number of layers and type of activation functions, random initialization of parameters along with a random shuffle of the datasets, and hyper-ensembles, in which ensembles with different hyperparameters are combined~\cite{he2023survey}.

In this paper, we rely on random initialization of deep ensembles, which has shown promising results for many practical problems~\cite{he2023survey,fort2019deep}. \autoref{fig:de} provides a visual representation of this method. DE is a mixture model:

\begin{equation*}
    \centering
    p(y|x) = \frac{1}{N} \sum_{n=1}^{N} p^{\ast}_n(y|x, \omega_n)
\end{equation*}

where the predictions are combined into one output $\mu$ (interpreted as a mixture of Gaussian distributions) and the variance of the outputs ($\sigma$ in \autoref{fig:mc-dropout}) measures the uncertainty.  

Deep Ensembles provide a robust measure of uncertainty that is able to account for multiple sources of model and data uncertainty~\cite{he2023survey}.
For DE, the main hyperparameter is the number of models ($N$). For large values of $N$, DE provides a precise implementation of the BNN approach, a theoretically grounded approach that provides the best uncertainty quantification while, however, being associated with a high computational cost. 
Thus, the trade-off between the precision of the BNN approximation and computational cost must be assessed in each application domain, such as ADS. An advantage of DE is that it is widely applicable, as it does not require any modification of any existing DNN. However, the computational overhead associated with training multiple models and loading them simultaneously in memory during inference might be unacceptable for a large number of models.
In this paper, we aim to assess the effectiveness and performance of DE as a failure predictor for ADS testing under a large number of ensemble sizes.

\subsection{Implementation}\label{sec:implementation}

We implemented our codebase in Python and made it publicly available~\cite{replication-package}. We support ADS models written in Tensorflow/Keras integrated in the Udacity simulator for self-driving cars~\cite{udacity-simulator}. Both UQ methods (\mcdropout and Deep Ensembles) were tested on instances of NVIDIA's \davetwo~\cite{nvidia-dave2} models. For \mcdropout, a dropout layer was added between each layer of the original model.

\section{Empirical Evaluation}\label{sec:evaluation}
\subsection{Research Questions}

We consider the following research questions:

\noindent
\textbf{RQ\textsubscript{1} (effectiveness):}
How effective is UQ at predicting failures of ADS? 
What is the best configuration in terms of dropout rate and number of samples (for MCD) or number of models (for DE)? How does the effectiveness vary when considering different confidence levels?

\noindent
\textbf{RQ\textsubscript{2} (prediction over time):}
How does the prediction power of UQ change when considering different detection periods?

\noindent
\textbf{RQ\textsubscript{3} (comparison):}
How does UQ  compare with \selforacle~\cite{2020-Stocco-ICSE} and \ty~\cite{2022-Stocco-ASE} in terms of effectiveness?

\noindent
\textbf{RQ\textsubscript{4} (performance):}
What is the performance of running UQ in terms of time overhead in making predictions? How do the UQ methods compare with \selforacle and \ty?

The first research question (RQ\textsubscript{1}) aims to assess whether our approach is able to attain a high failure prediction rate and which method (i.e., \mcdropout, Deep Ensembles, and their parameters) yields the best prediction score. 
Failure prediction is only useful if it helps to anticipate a failure, which is studied in the second research question (RQ\textsubscript{2}). 
To assess the usefulness of UQ methods over existing solutions, the third research question (RQ\textsubscript{3}) compares UQ with two state-of-the-art failure predictors for ADS~\cite{2020-Stocco-ICSE,2022-Stocco-ASE}.
The last research question (RQ\textsubscript{4}) evaluates the runtime cost of each technique, to assess efficiency in conjunction with effectiveness.

\subsection{Experimental Setup}

In this paper, we follow the same experimental setting of the original papers we compare against~\cite{2020-Stocco-ICSE,2022-Stocco-ASE}, in terms of simulation platform, objects of study, and metrics. We briefly summarize the experimental setup next.

\subsubsection{ADS Under Test}

To implement DNN-based ADS, we use NVIDIA's \davetwo model~\cite{nvidia-dave2}, a reference model widely used as the object of study in prior related work~\cite{deepxplore,deeptest,deeproad,2020-Stocco-ICSE,2020-Riccio-EMSE,2021-Jahangirova-ICST,deepcrime,2024-Lambertenghi-ICST}.
\davetwo consists of three 5x5 convolutional layers with stride 2 plus two 3x3 convolutional layers (no stride applied), followed by five fully-connected layers with a dropout rate of $0.05$ and ReLu activation function. For the experiments with \selforacle and \ty, we obtained the trained \davetwo models from the replication package of our baselines~\cite{2020-Stocco-ICSE,2022-Stocco-ASE}, to make sure to test the same ADS used in the previous work. For UQ, we had no choice but to retrain \davetwo (details available in \autoref{sec:tool-configuration}).

\subsubsection{Driving Simulator}

We tested UQ  through simulation-based testing, which is the standard practice for testing ADS and their behaviour prior to real-world deployment~\cite{2020-Haq-ICST,12233,2022-Stocco-TSE,2023-Stocco-EMSE}. We simulate the ADS testing practices customary of industry, where testers use a closed-loop track in a virtual environment, prior to on-road testing on public roads~\cite{Cerf:2018:CSC:3181977.3177753,10-million-miles,waymo-driver,waymos-secret-testing}.
While our approach is independent from the chosen simulation platform, in our study to test the lane-keeping ADS we used the Udacity simulator for self-driving cars~\cite{udacity-simulator}, a cross-platform driving simulator developed with Unity3D~\cite{unity}, used in the ADS testing literature~\cite{2020-Stocco-ICSE,2020-Riccio-EMSE,2021-Jahangirova-ICST,knoll-monitoring,2020-Stocco-GAUSS}, including our baselines~\cite{2020-Stocco-ICSE,2022-Stocco-ASE}. The simulator supports various closed-loop tracks for testing behavioural cloning ADS models, as well as the ability to generate changeable environmental perturbations (e.g., weather effects), which is useful to test an ADS on both nominal and unseen conditions. We chose the default sunny weather condition as the reference nominal scenario.

\subsubsection{Benchmark}

Concerning our evaluation set, we consider three existing datasets of simulations from previous work~\cite{2022-Stocco-ASE}. The first two datasets deal with failures induced by \textit{out-of-distribution conditions} (OOD). An ADS that has been trained on some given nominal conditions and environment can fail in different instances of that environment. The first OOD benchmark (OOD\textsubscript{extreme}) is characterized by severe illumination/weather conditions with respect to the nominal sunny scenario (see \autoref{fig:conditions}). These conditions are available from the replication package of the \selforacle paper~\cite{2020-Stocco-ICSE} and account for 7 simulations with different degrees of extreme OOD conditions: day/night, rain, snow, fog, day/night + rain, day/night + snow, day/night + fog. 
The second OOD benchmark (OOD\textsubscript{moderate}) consists of milder weather conditions without the strong luminosity changes present in the OOD\textsubscript{extreme} benchmark. 
Overall, concerning the OOD benchmarks, a total of 51 OOD one-lap simulations were collected: 21 for OOD\textsubscript{extreme} and 30 for OOD\textsubscript{moderate} (10 $\times$ rain, 10 $\times$ fog, 10 $\times$ snow). 
The third benchmark (Mutants) consists of faulty ADS models produced by mutation testing~\cite{deepcrime}. In this case, the ADS drives under nominal  (sunny) conditions, but it can occasionally fail due to inadequate training, a frequent scenario during the development process of an ADS model (i.e., data collection, training, and testing is an iterative process~\cite{2020-Riccio-EMSE}). 
Overall, the evaluation set comprises 265 failures that our approach is expected to detect timely. 
Both scenarios are of interest to our work, as a failure predictor should be agnostic about the conditions that cause the failures (i.e., unknown inputs or DNN model bugs). 
Moreover, to estimate the threshold used by UQ methods, the evaluation set includes simulations under nominal sunny weather conditions (one for each of three benchmarks OOD\textsubscript{extreme}, OOD\textsubscript{moderate}, and Mutants) using the robust, unmutated, \davetwo model.  

\subsubsection{Detection Windows in Evaluation Set}\label{sec:labelling}\label{sec:detection-windows}
The Udacity simulator automatically labels individual failing frames as either nominal or failing, according to whether the ADS was on track or off-track, respectively. We focus on the part of the simulation \textit{preceding each failure}, whereas the frames labeled as failing are not considered. When a simulation exhibits multiple failures, we assess each failure individually. Differently from the compared papers~\cite{2020-Stocco-ICSE,2022-Stocco-ASE}, for all benchmarks, we calculate the actual frame rate of each simulation, instead of using a fixed window size of 15 frames. This choice was motivated by the fact the three benchmarks were captured on different machines and hardware, at different frame rates. Consequently, using a fixed window size would fail to uniformly represent simulation time across all datasets, making it challenging to fairly evaluate the performance of our predictors.

\subsubsection{Baselines}

As described in \autoref{sec:existing-unsup-failure-prediction}, we use two baselines for UQ. 
Concerning \selforacle we consider the best configuration presented in the original paper, which uses a variational autoencoder~\cite{An2015VariationalAB} (VAE) with a latent size of 2, trained to minimize the MSE (see \autoref{sec:existing-unsup-failure-prediction}) between the original and reconstructed nominal images (sunny). Regarding \ty, we assessed the best configuration that includes heatmap derivative as a summarization method. 

\subsubsection{Configurations}\label{sec:tool-configuration}

For both UQ methods, we trained lane-stable \davetwo models using an existing dataset ~\cite{2020-Stocco-ICSE} with more than $32k$ images on nominal sunny conditions following two different track orientations (normal, reverse), and additional data for recovery. Each image is labeled with the human expert-provided ground truth steering angle value for that driving image. The maximum driving speed of the driving model was 30 mph during data generation, the default value in the Udacity simulator. 

For \mcdropout, we trained several \davetwo models varying two parameters. The first parameter is the dropout rate, which we vary in the range $[0.05, 0.1, 0.15, 0.20, 0.25, 0.30, 0.35]$. Models with a dropout rate higher than $0.40$ were disregarded for not being able to complete a lap in the simulator. The second parameter is the number of samples, which we vary in the range $[2, 3, 4, 5, 10, 20, 32, 64, 128]$. For Deep Ensembles, we trained several \davetwo models varying the number of models in the ensemble, considering the range $[2, 3, 4, 5, 6, 7, 10, 30, 50, 70, 90, 100, 120]$.

The number of epochs was set to 50, with a batch size of 128 and a learning rate of 0.0001. We used early stopping with a patience of 10 and a minimum loss change of 0.0005 on the validation set. The network uses the Adam optimizer~\cite{Kingma2014AdamAM} to minimize the MSE between the predicted steering angles and the ground truth value. We used data augmentation to mitigate the lack of image diversity in the training data. Specifically, 60\% of the data was augmented through different image transformation techniques (e.g., flipping, translation, shadowing, brightness). We cropped the images to 80x160 and converted them from RGB to YUV color space. We only retained solid models for testing, i.e., models able to drive multiple laps in each track under nominal conditions without showing any misbehavior in terms of crashes or out-of-track events. This should also provide more guarantees about the quality of the uncertainty score estimations obtained from white box access to the models. 

Overall, our experiment includes 232 models under test. For \mcdropout, we trained 63 final models (7 dropout rates $\times$ 9 number of samples) for parameter optimization and did further testing on the best dropout rates to study the distribution. For Deep Ensembles, we trained 138 different models and built 30 different ensembles. For smaller-sized ensembles [2-5] we tested various combinations of models to study their effectiveness. As our evaluation set comprises 380,717 images, overall we computed 15,723,347 uncertainty scores in our experiments (11.5 days computing time). 


\begin{table*}[t]


\caption{RQ\textsubscript{1-2-3}: Results for the best failure predictors. Bold = average $F_{3}$ scores; grey = best $F_{3}$ scores.\label{tab:rq1}}

\setlength{\tabcolsep}{2.6pt}
\renewcommand{\arraystretch}{1.1}

\begin{tabular}{lclll@{\hskip 1.5em}lll@{\hskip 1.5em}lll@{\hskip 1.5em}lll@{\hskip 1.5em}lll@{\hskip 1.5em}lll@{\hskip 1.5em}lll@{\hskip 1.5em}lll}

\toprule

\multirow{2}{*}{} &  & \multicolumn{3}{c}{MCD5 S32} & \multicolumn{3}{c}{MCD5 S64} & \multicolumn{3}{c}{MCD5 S128} & \multicolumn{3}{c}{DE5} & \multicolumn{3}{c}{DE10} & \multicolumn{3}{c}{DE50} & \multicolumn{3}{c}{SelfOracle} & \multicolumn{3}{c}{ThirdEye} \\
&  & \multicolumn{3}{c}{conf = 0.99} & \multicolumn{3}{c}{conf = 0.99} & \multicolumn{3}{c}{conf = 0.99} & \multicolumn{3}{c}{conf = 0.999} & \multicolumn{3}{c}{conf = 0.999} & \multicolumn{3}{c}{conf = 0.999} & \multicolumn{3}{c}{conf = 0.99} & \multicolumn{3}{c}{conf = 0.95} \\
 
\cmidrule(r){3-5}
\cmidrule(r){6-8}
\cmidrule(r){9-11}
\cmidrule(r){12-14}
\cmidrule(r){15-17}
\cmidrule(r){18-20}
\cmidrule(r){21-23}
\cmidrule(r){24-26} 

& TTF (s) & Pr & Re & $F_{3}$ & Pr & Re & $F_{3}$ & Pr & Re & $F_{3}$ & Pr & Re & $F_{3}$ & Pr & Re & $F_{3}$ & Pr & Re & $F_{3}$ & Pr & Re & $F_{3}$ & Pr & Re & $F_{3}$ \\ \midrule
\multirow{4}{*}{\quad \bf OOD\textsubscript{extreme}} 
 & 1 & 22 & 93 & 69 & 19 & 100 & 69 & 22 & 93 & 69 & 42 & 100 & 87 & 42 & 100 & 87 & 100 & 100 & 100 & 73 & 100 & 96 & 19 & 93 & 65 \\
 & 2 & 23 & 100 & 73 & 19 & 95 & 66 & 20 & 88 & 65 & 42 & 100 & 87 & 42 & 100 & 87 & 100 & 100 & 100 & 73 & 96 & 93 & 19 & 95 & 66 \\ 
 & 3 & 23 & 96 & 71 & 17 & 82 & 59 & 22 & 89 & 67 & 43 & 100 & 87 & 43 & 100 & 87 & 100 & 100 & 100 & 70 & 89 & 86 & 19 & 93 & 66 \\ 
 & avg & 22 & 96 & \textbf{71} & 18 & 92 & \textbf{65} & 21 & 90 & \textbf{67} & 42 & 100 & \textbf{87} & 42 & 100 & \textbf{87} & 100 & 100 & \textbf{100} & 72 & 95 & \textbf{92} & 19 & 94 & \textbf{66} \\  [0.5em]

\multirow{4}{*}{\quad \bf OOD\textsubscript{moderate}} 
& 1 & 31 & 100 & 80 & 30 & 98 & 79 & 30 & 98 & 79 & 100 & 100 & 100 & 100 & 100 & 100 & 100 & 100 & 100 & 51 & 98 & 89 & 13 & 87 & 54 \\ 
& 2 & 27 & 86 & 69 & 26 & 83 & 67 & 25 & 81 & 65 & 100 & 97 & 97 & 100 & 98 & 98 & 100 & 100 & 100 & 47 & 91 & 83 & 11 & 75 & 47 \\ 
& 3 & 21 & 63 & 51 & 21 & 63 & 51 & 20 & 63 & 51 & 72 & 70 & 70 & 89 & 79 & 79 & 72 & 70 & 70 & 33 & 62 & 57 & 10 & 62 & 40 \\ 
& avg & 26 & 83 & \textbf{67} & 26 & 81 & \textbf{66} & 25 & 81 & \textbf{65} & 91 & 89 & \textbf{89} & 96 & 92 & \textbf{93} & 91 & 90 & \textbf{90} & 44 & 84 & \textbf{76} & 12 & 75 & \textbf{47} \\ [0.5em]

\multirow{4}{*}{\quad \bf Mutants} 
& 1 & 65 & 100 & 94 & 65 & 99 & 94 & 65 & 100 & 94 & 100 & 100 & 100 & 100 & 100 & 100 & 100 & 100 & 100 & 77 & 82 & 81 & 44 & 99 & 87 \\ 
& 2 & 65 & 98 & 93 & 64 & 97 & 92 & 64 & 97 & 92 & 100 & 96 & 96 & 100 & 97 & 97 & 100 & 97 & 97 & 61 & 49 & 50 & 44 & 97 & 86 \\ 
& 3 & 60 & 88 & 84 & 59 & 85 & 81 & 59 & 87 & 83 & 100 & 81 & 82 & 100 & 87 & 87 & 94 & 81 & 82 & 56 & 41 & 41 & 41 & 91 & 80 \\ 
& avg & 63 & 95 & \textbf{90} & 63 & 94 & \textbf{89} & 63 & 95 & \textbf{90} & 100 & 92 & \textbf{93} & 100 & 95 & \textbf{95} & 98 & 93 & \textbf{93} & 65 & 57 & \textbf{57} & 43 & 95 & \textbf{84} \\
\cmidrule(r){1-26} 

\multirow{4}{*}{\it Average (All) }
& 1 & 39 & 98 & 81 & 38 & 99 & 81 & 39 & 97 & 81 & 81 & 100 & 96 & 81 & 100 & 96 & 100 & 100 & 100 & 67 & 94 & 89 & 25 & 93 & 69 \\
& 2 & 38 & 95 & 79 & 36 & 92 & 75 & 37 & 89 & 74 & 81 & 97 & 93 & 81 & 98 & 94 & 100 & 99 & 99 & 60 & 79 & 75 & 25 & 89 & 66 \\ 
& 3 & 34 & 83 & 69 & 32 & 77 & 64 & 34 & 80 & 67 & 71 & 84 & 80 & 77 & 88 & 85 & 89 & 84 & 84 & 53 & 64 & 61 & 24 & 82 & 62 \\ 
& avg & 37 & 92 & \textbf{\cellcolor{lightgray}76} & 36 & 89 & \textbf{73} & 36 & 88 & \textbf{74} & 78 & 94 & \textbf{90} & 79 & 96 & \textbf{92} & 96 & 94 & \cellcolor{lightgray}\textbf{94} & 60 & 79 & \cellcolor{lightgray}\textbf{75} & 24 & 88 & \cellcolor{lightgray}\textbf{66} \\

\bottomrule

\end{tabular}
\end{table*}

\subsubsection{Metrics used for Analysis}\label{sec:metrics_analysis}

To answer RQ\textsubscript{1}, RQ\textsubscript{2}, and RQ\textsubscript{3}, we apply a window function on non-overlapping, fixed length, sequences of scores, returning the \textit{maximum} score within a window. In previous work~\cite{2022-Stocco-ASE}, the \textit{arithmetic mean} of the scores within a window was also used, with less promising results. Therefore, in this paper, we limit our investigation to the maximum window function.
The sets of (windowed) uncertainty confidence scores represent a model of normality collected in nominal driving conditions using different methods for computing the uncertainty profiles. Following existing literature~\cite{2022-Stocco-ASE,2020-Stocco-ICSE}, we use probability distribution fitting to obtain a statistical model of the uncertainty scores. We set a threshold $\gamma$ for the expected false alarm rate in nominal conditions and estimate the shape $\kappa$ and scale $\theta$  of a fitted Gamma distribution of the uncertainty scores to ensure the expected false alarm rate is below the chosen threshold $\gamma$~\cite{2020-Stocco-ICSE}. 
In this study, we experiment with different thresholds, varying  $\gamma$ in the range $[0.95, 0.99, 0.999, 0.9999, 0.99999]$, hence expanding substantially the $\gamma$ threshold ranges considered previously (\ty was only evaluated for $\gamma=0.95$, whereas \selforacle was evaluated for $\gamma=0.95$ and $\gamma=0.99$).

We compute the true positives as the number of correct failure predictions within a detection window and the false negatives as the number of missed failure predictions when our framework does not trigger an alarm in a detection window. The false positives and true negatives are measured using nominal simulations to which analogous windowing is applied. 
Our primary goal is to achieve a high Recall (Re), or true positive rate, defined as Re=TP/(TP+FN)). Recall measures the fraction of safety-critical failures detected by a technique. It is also important to achieve high precision (Pr), defined as Pr=TP/(TP+FP). Precision measures the fraction of correct warnings reported by a technique.
Consistent with previous work~\cite{2022-Stocco-ASE}, we consider the $F_{beta}$ score~\cite{https://doi.org/10.1002/asi.4630300621}, with $\beta=3.0$, as a weighted balance between precision and recall ($F_{3}=\frac{10 \cdot \text{Precision} \times \text{Recall}}{9 \cdot \text{Precision} + \text{Recall}}$), staying consistent with previous work. 
We are interested in an F-measure that weights recall higher compared to precision because the cost associated with false negatives is very high in the safety-critical domain~\cite{https://doi.org/10.1002/asi.4630300621                                  } as it means a missed failure detection. In contrast, in our setting, the cost associated with false positives (false alarms) is relatively lower compared to false negatives. 
We also compute the threshold-independent metric AUC-ROC (area under the curve of the Receiver Operating Characteristics), which we use to choose the top three models as presenting the results for all models would be infeasible. 
For RQ\textsubscript{2}, for each failure, we adopt a detection window granularity equal to one second of simulation in the Udacity simulator and we consider window sizes from 1 to 3 seconds prior to the failures (time to failure, TTF for short).
Previous studies in the Udacity simulator~\cite{2021-Stocco-JSEP} indicate a TTF of 3 seconds as sufficient to avoid failures at 30 mph, which is the constant cruising speed of the ADS in the simulator.

To answer RQ\textsubscript{4}, we compute the execution time (in milliseconds \textit{ms}) and RAM usage during inference using the Python tool \texttt{mprofile}~\cite{mprofile} on a machine featuring an AMD Ryzen 7 3800XT 8-Core (16 Threads) Processor, 32GB system RAM and a NVIDIA 3070 GPU with 8GB of VRAM. All models were evaluated using two laps under normal conditions for a total evaluation set consisting of 11,031 images. All inferences were computed using the CPU only with all 16 (virtual) cores enabled. For Deep Ensembles, all models of an ensemble were loaded into memory and performed the inference concurrently. For \mcdropout, the model was loaded into memory and performed the inference concurrently, running the inference process multiple times as multiple parallel threads. For \selforacle and \ty, the cache was cleared, forcing the models to compute the heatmaps during inference instead of relying on pre-computed values. 

\subsection{Results}

\begin{figure}[t]
\centering
\includegraphics[trim= 1.5cm 0cm 2cm 2cm, clip=true, width=\linewidth]{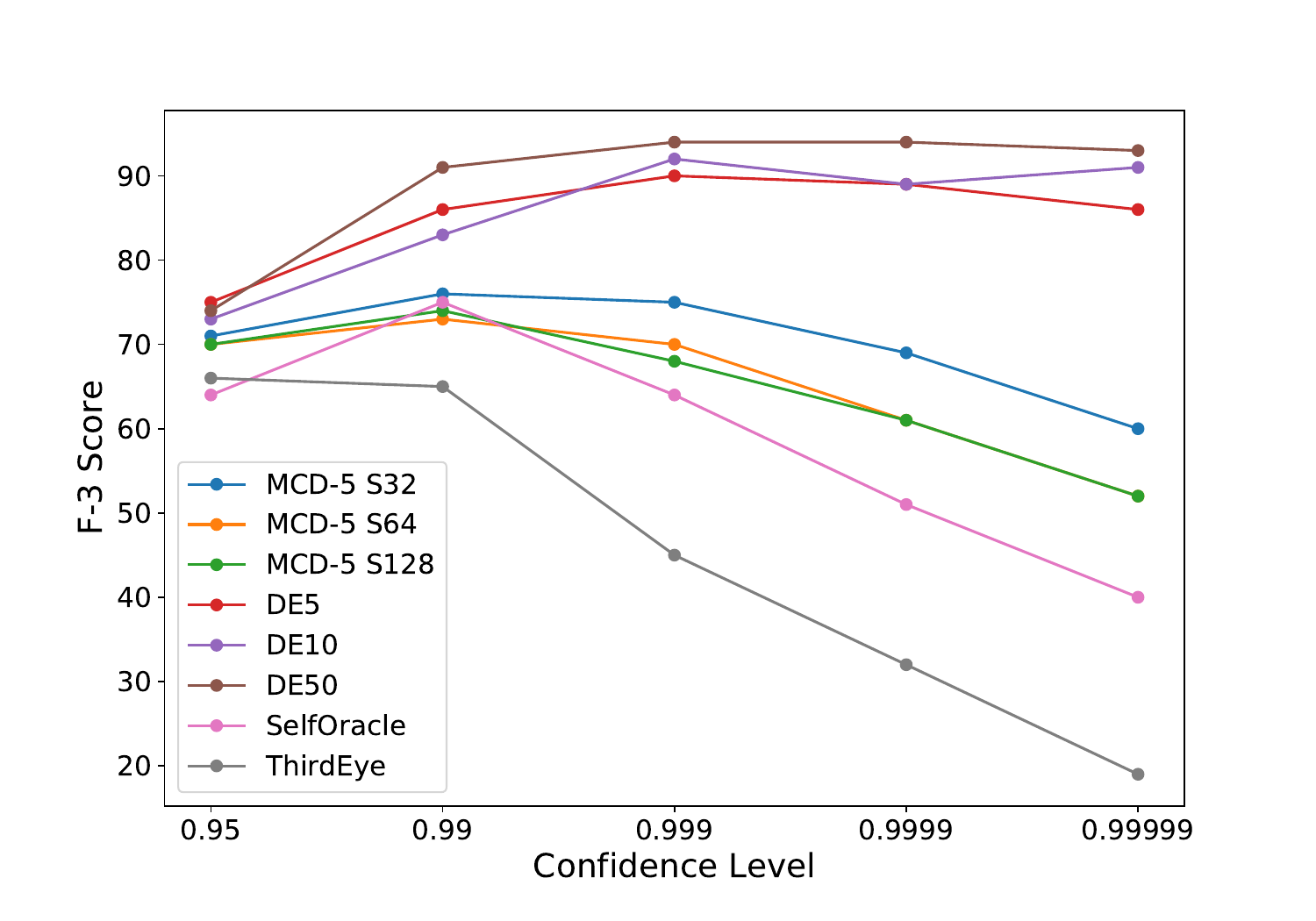}
\caption{RQ\textsubscript{1}: $F_{3}$ scores for the best failure predictors across all confidence levels.}
\label{fig:f3}
\end{figure}

\subsubsection{Effectiveness (RQ\textsubscript{1})}
For MCD, the top three configurations from our experiments are MCD models with dropout rate=0.05 and number of samples 32, 64, and 128. In the rest of the paper, we refer to them as MCD-5 S32, MCD-5 S64, and MCD-5 S128, respectively.
For DE, the top three configurations from our experiments are DE with 5, 10, and 128 models, referred to as DE5, DE10, and DE50 next.
\autoref{fig:f3} reports the three models for each UQ method and the two baselines at different confidence values. Deep Ensembles models perform well across all confidence levels. \mcdropout models perform well at confidence levels $\gamma=0.95$ and $\gamma=0.99$ and worse with higher confidence levels. Consequently, in the rest of the paper, we report detailed results considering the optimal confidence threshold for each model.

\autoref{tab:rq1} presents the effectiveness results for the top three configurations of UQ (\mcdropout, Deep Ensembles), \selforacle, and \ty. Results are averaged across conditions, split between external unknown conditions (OOD\textsubscript{extreme} and OOD\textsubscript{moderate}) and internal uncertain conditions (Mutants). For each condition, we evaluate failure detection with a detection window of 1-3 seconds and also report the average of these scores.
The effectiveness metrics consider the optimal confidence threshold for each model (\autoref{fig:f3}). 
Precision (Pr) is measured in anomalous conditions, which explains why it is lower than the expected value associated with the confidence threshold in most cases. Due to space constraints, in this section, we only comment on the average $F_{3}$ scores over all benchmarks.
On average, UQ with \mcdropout reaches a $F_3$ score of $73-76$c. UQ with Deep Ensembles, on the other hand, performs better with $F_3$ scores of $90-94$. For the MCD-5 model, increasing the sample size does not improve the effectiveness but rather causes a slight drop in the $F_3$ score. The precision for MCD-5 remains relatively low across all sample sizes larger than 32, indicating that the false positive rate does not improve with a higher number of samples. For Deep Ensembles, the theoretical best performance is DE50 (i.e., an ensemble of 50 models) with an $F_3$ of $94$, outperforming any other configuration. In practice, though, a DE of 50 models might be computationally expensive, therefore DE5 or DE10 are more likely to be used. All Deep Ensembles models have a high recall and a low false positive rate (i.e., high precision).

\begin{tcolorbox}
\begin{center}
\begin{minipage}[t]{0.99\linewidth}
\textbf{RQ\textsubscript{1}}: \textit{
UQ with Deep Ensembles (5/50 models) is the best-performing failure predictor for ADS, achieving the highest failure prediction rates across all conditions ($F_3$ = 90-94\%).
}
\end{minipage}
\end{center}
\end{tcolorbox}

\subsubsection{Prediction Over Time (RQ\textsubscript{2})}
\autoref{tab:rq1} reports the effectiveness considering different time to failure (TTF, Column~2). In principle, failure prediction should get more challenging as we move farther from the failure. This is confirmed for all configurations of UQ (considering the average scores) with the prediction power dropping ($F_{3}$) slightly when we move from a 1-second detection window to a 2-second window and a larger drop when considering a 3-second window. The best \mcdropout model performance drops by -3.5\% and -14.8\% at 2 and 3 seconds TTF respectively, compared to 1 second TTF. The best Deep Ensembles model performance drops by -1\% and -16\% respectively. When we look at the OOD\textsubscript{extreme} benchmark, we observe that Deep Ensembles of all sizes do not drop any predictive power up to 3 seconds TTF, with DE50 predicting all failures.

\begin{tcolorbox}
\begin{center}
\begin{minipage}[t]{0.99\linewidth}
\textbf{RQ\textsubscript{2}}: \textit{
On average, the effectiveness of the best configurations of UQ drops by 16\% average $F_3$ up to 3 seconds before the failures. The effectiveness of UQ with Deep Ensembles remains high under OOD\textsubscript{extreme} conditions (no decrease in $F_3$) up to 3 seconds before the failure.
}
\end{minipage}
\end{center}
\end{tcolorbox}

\subsubsection{Comparison (RQ\textsubscript{3})}
Considering the average $F_{3}$ scores across benchmarks from \autoref{tab:rq1}, the best configurations of both UQ methods are superior to \selforacle and \ty at predicting misbehaviours. \mcdropout with a 5\% dropout rate and 32 sample size is comparable to \selforacle with a +1\% improvement in $F_3$ score. Compared to \ty, MCD5-S32 is +15\% better at predicting misbehaviour ($F_3$). Deep Ensembles 50 outperforms both \selforacle and \ty, with an improvement of +25\% and +42\% in average $F_3$ scores, respectively. 
On the OOD\textsubscript{extreme} benchmark, UQ  scores a +8.7\% and +51\% increase in $F_{3}$ w.r.t. \selforacle and \ty. 
For OOD\textsubscript{moderate} conditions, average $F_{3}$ scores raise up to 93, for DE10, whereas the best $F_3$ from our baseline (\selforacle) is 76\%. 
For Mutants, our results show a remarkable difference in effectiveness between UQ over \selforacle and \ty. Both \mcdropout and Deep Ensembles score higher with average $F_3$ scores in the range of 89-95,  a +66.7\% w.r.t. \selforacle in $F_3$ and 13\% w.r.t. \ty. 

Overall, average results for $F_{3}$ show significant improvements of UQ  over previous experiments. For Deep Ensembles, this finding holds independent of the configuration being used and the reaction period considered.

We assessed the statistical significance of these differences using the non-parametric Mann-Whitney U test~\cite{Wilcoxon1945} (with $\alpha = 0.05$) and the magnitude of the differences using the Cohen's $d$ effect size~\cite{cohen1988statistical}. The difference in $F_{3}$ score between Deep Ensembles and \selforacle and \ty were found to be statistically significant ($p$-value $<$ 0.05) with medium and large effect sizes. As expected by looking at the average $F_{3}$ scores of \autoref{tab:rq1}, there is no statistically significant difference between MCD and \selforacle ($p$-value $\geq$ 0.05), whereas the difference with \ty is statistically significant with medium effect size.

\begin{tcolorbox}
\begin{center}
\begin{minipage}[t]{0.99\linewidth}
\textbf{RQ\textsubscript{3}}: \textit{
UQ with Deep Ensembles outperforms \selforacle and \ty in terms of failure prediction under all conditions, with statistical significance.
}
\end{minipage}
\end{center}
\end{tcolorbox}

\begin{figure}[b]
\centering
\includegraphics[trim=1cm 0cm 2.45cm 2cm, clip=true, width=0.95\linewidth]{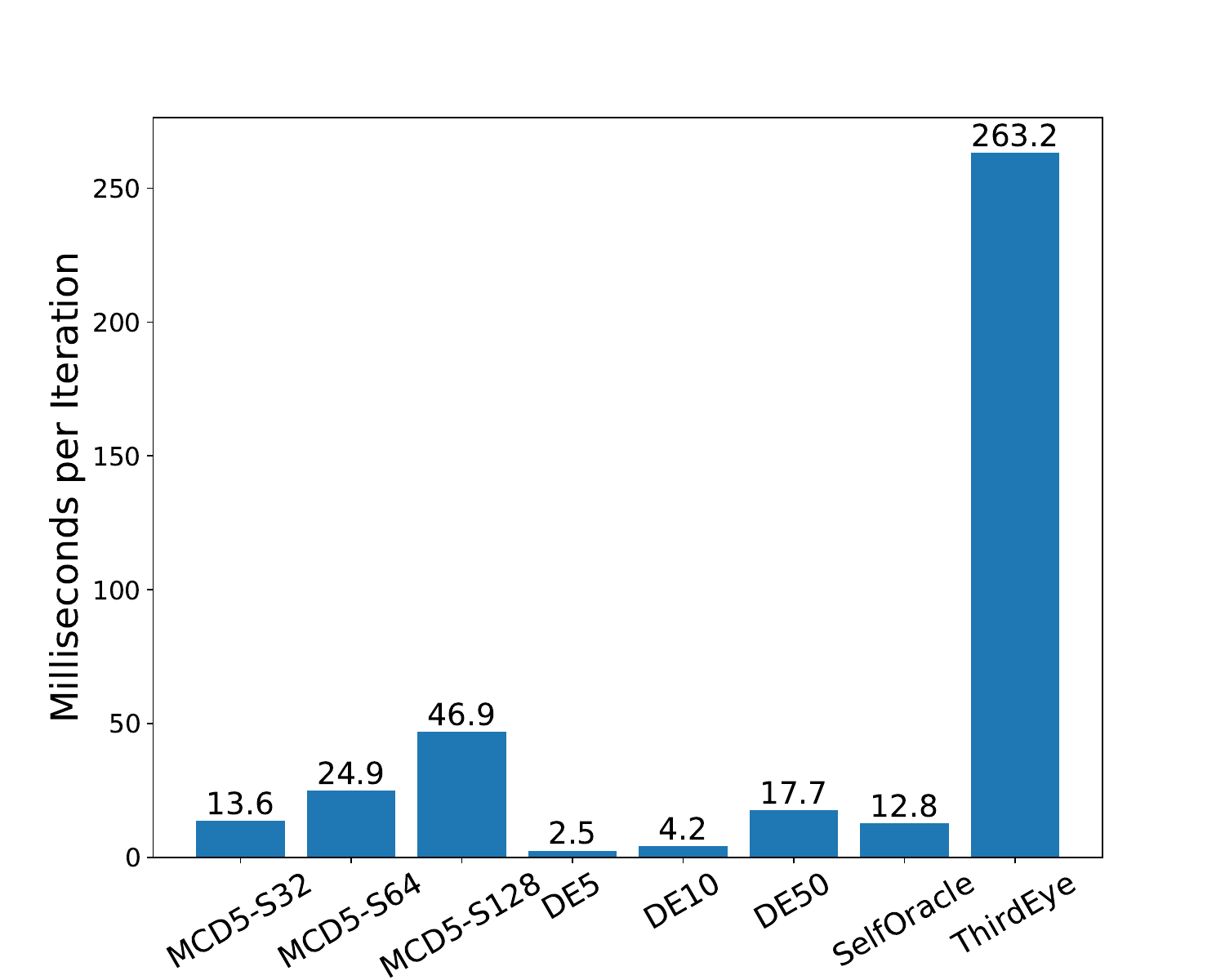}
\caption{RQ\textsubscript{4}: Computational overhead (ms/iteration).}
\label{fig:rq4}
\end{figure}

\subsubsection{Performance (RQ\textsubscript{4})}

\autoref{fig:rq4} shows the results of the different models in ms per iteration/image.
Deep Ensembles performed best with the DE5 employing 2.5 ms/image and the DE50 employing 17.7 ms/image. 
\selforacle has a similar performance to larger deep ensembles with 12.8 ms/image. \mcdropout performs worse than deep ensembles and \selforacle. MCD-5 S32 took 13.6 ms/image and 46.9 ms/image for MCD-5 S128. Both UQ methods, as expected, decrease in performance when either the number of models or the sample size increases. \ty, as seen in \autoref{fig:rq4}, takes significantly longer than any other method to process an image, being 105$\times$ \textit{slower} than Deep Ensembles. This performance is expected, as \ty needs to compute the heatmap for each image, which is a computationally expensive process.

Concerning the memory usage of the different models, we do not report extensive results, but we discuss a few interesting insights. \mcdropout used the least amount of memory considering its best configuration (635 MB). Deep Ensembles 50 used a larger amount of memory, requiring loading all models into memory simultaneously (1.37 GB). The size of each model itself (used in \mcdropout or the Deep Ensembles) is approximately 4.7 MB. \selforacle and \ty used the most amount of memory, requiring 27.6 GB and 7.3 GB of memory respectively.

\begin{tcolorbox}
\begin{center}
\begin{minipage}[t]{0.99\linewidth}
\textbf{RQ\textsubscript{4}}: \textit{
Small Deep Ensembles are the most computationally efficient outperforming \ty and \selforacle. Particularly, DE5 and DE10 employ on average less than 5 ms/image.
}
\end{minipage}
\end{center}
\end{tcolorbox}

\subsection{Threats to Validity}\label{sec:ttv}

\subsubsection{Internal validity}
All variants of UQ, \selforacle, and \ty were compared under identical experimental settings and on the same evaluation set. Thus, the main threat to internal validity concerns our implementation of the testing scripts to evaluate the failure prediction scores, which we tested thoroughly. 
Concerning the training of the ADS model, we used artifacts publicly available in the replication packages of the \selforacle~\cite{2020-Stocco-ICSE} and \ty~\cite{2022-Stocco-ASE} papers. Regarding the simulation platform, to allow a fair comparison, we used the Udacity simulator adopted in analogous failure prediction studies~\cite{2020-Stocco-ICSE,2022-Stocco-ASE}. However, it is important to note our approach is independent of the chosen simulation platform. Other open-source propositions are available, such as CARLA~\cite{carla}, LGSVL~\cite{rong2020lgsvl}, and BeamNG~\cite{beamng}. CARLA and LGSVL mostly deal with urban environments with static and dynamic obstacles, whereas BeamNG is conceptually similar to Udacity as it was used in similar lane-keeping testing studies~\cite{2020-Riccio-FSE,biagiola2023better,Gambi:2019:ATS:3293882.3330566}. We discard commercial close-source solutions such as Siemens PreScan~\cite{prescan}, ESI Pro-SiVIC~\cite{pro-sivic}, and PTV VISSIM~\cite{vissim} as they do not allow full replicability of our results and also focus on urban scenarios or other ADS tasks such as automated valet parking or breaking assistance. 

\subsubsection{External validity}
The limited number of self-driving systems in our evaluation constitutes a threat to the generalizability of our results to other ADS. Moreover, results may not generalize, or generalize differently, when considering other simulation platforms than Udacity. For the uncertainty scores, we considered two quantification methods, and the effectiveness of our tool may change when considering different strategies. To mitigate these issues, we selected the most popular techniques for computing uncertainties in regression deep neural networks, as outlined in Weiss and Tonella~\cite{Weiss2021FailSafe}. 

\subsubsection{Reproducibility} 
All our results, the source code, and the simulator are accessible and can be reproduced~\cite{replication-package}.

\section{Discussion}\label{sec:discussion}

\subsection{UQ for Failure Prediction}

Our research emphasizes the intricate nature and diverse range of failure scenarios that runtime monitoring techniques must address. Uncertainty scores, usually employed quantitatively by humans to understand deep neural network mispredictions, were used in this study as a cumulative error scoring function over time. This approach assumes that these scores contain valuable information for assessing the behavior of DNNs and, by extension, of the autonomous driving systems that rely on them.

Our approach relies on the efficacy of uncertainty scores as a technique for assessing the nominal driving behavior of ADS. A well-trained DNN would excel in capturing relevant structures in an image, such as road lanes, resulting in more precise uncertainty scores compared to inadequately trained DNNs. Furthermore, methods for quantifying uncertainty provide a more transparent and efficient means of evaluating ADS behavior than opaque data- or black-box techniques. Our findings confirm that UQ methods outperform existing techniques in both out-of-distribution and mutation testing scenarios.

\subsection{Discussing UQ Configurations}

In our benchmarks, UQ using \mcdropout exhibited superior performance on the Mutants dataset compared to both OOD\textsubscript{moderate} and OOD\textsubscript{extreme}. It demonstrated the capability to predict 95\% failures with an acceptable precision, up to 3 seconds in advance. This observation underscores the effectiveness of \mcdropout as a reliable metric for understanding internal model uncertainty. Conversely, UQ with Deep Ensembles consistently delivered remarkable prediction results across all benchmarks. Even for different confidence levels (\autoref{fig:f3}), Deep Ensembles consistently outperformed alternative methods. Our findings confirm that Deep Ensembles excels at capturing uncertainty from diverse sources and outshines \mcdropout~\cite{he2023survey}. 

While the theoretical best ensemble with 50 models may not be practical for real-world applications in ADS, our Deep Ensembles with only 5 models outperformed all other techniques and exhibited robust uncertainty estimates, with performance similar to DE50. Taking into account computational runtime, we found that smaller Deep Ensembles were more efficient than \mcdropout. This advantage stems from the ability to load and run multiple models concurrently, provided the hardware can support the model sizes. In contrast, \mcdropout requires less memory but still needs to run the inference multiple times (32-128), making it less competitive than DE5 or DE10. Furthermore, implementing \mcdropout necessitates modifications to the ADS model. Considering all these factors, our comprehensive evaluation identifies UQ with Deep Ensembles as the optimal configuration, delivering the best results in our study.

\subsection{Comparison with Other Approaches}

As a baseline for our experiment, we used \selforacle and \ty from previous literature. In contrast to the previous experiment, we modified the evaluation as described in \autoref{sec:labelling} by implementing a dynamic window calculation for the OOD\textsubscript{extreme} benchmark. This allowed us to compare the benchmark scores more objectively. However, this caused the magnitude of the results for \selforacle and \ty to change~\cite{2020-Stocco-ICSE,2022-Stocco-ASE}. 
UQ with Deep Ensembles is a clear improvement over the baselines in terms of effectiveness and computation time. While each of the two baselines performed well in a specific benchmark (\selforacle in OOD\textsubscript{extreme} and \ty in Mutants), UQ with Deep Ensembles performs well across all benchmarks. Notably, even hybrid approaches with MCD + \selforacle or MCD + \ty are not expected to achieve higher scores than DE as they require more computational resources than Deep Ensembles. 
\section{Related Work}\label{sec:relwork}

\subsection{Anomaly Detection in Autonomous Driving}

We already discussed \selforacle~\cite{2020-Stocco-ICSE} and \ty~\cite{2022-Stocco-ASE}, for which we performed an explicit empirical comparison in this work. Similarly to \selforacle, DeepGuard~\cite{DeepGuard} uses the reconstruction error by VAEs to prevent collisions of vehicles with the roadside. 
DeepRoad~\cite{deeproad} uses embeddings created from features extracted by VGGNet~\cite{VGGNet} to validate driving images based on the distance to the training set.  
In other works~\cite{2021-Stocco-JSEP,2020-Stocco-GAUSS}, continual learning is used to minimize the false positives of a black-box failure predictor. 
Hell et al.~\cite{knoll-monitoring} evaluate VAEs, Likelihood Regret, and the generative modelling SSD, for ADS testing on OOD detection in the CARLA simulator. 
Michelmore et al.~\cite{MichelmoreWLCGK20,Michelmore} use Bayesian inference methods for probabilistic safety estimation.  
Henriksson et al.~\cite{Henriksson} use the negative of the log-likelihood as a black-box anomaly score. 
Borg et al.~\cite{arxiv.2204.07874} propose to pair OOD detection with VAEs with object detection for an automated emergency braking system. 
Strickland et al.~\cite{Strickland-ICRA-2018} use an LSTM solution with multiple metrics to predict collisions with vehicles at crossroads. 
Ayerdi et al.~\cite{ayerdi2023metamorphic} propose the use of metamorphic oracles to supervise a DNN-based ADS.  

Our approach reports extensive simulation-based testing results for both the effectiveness and efficiency of uncertainty quantification methods. For a broad overview of anomaly detection techniques in autonomous driving, we refer the reader to the survey by Bogdoll et. al~\cite{Bogdoll}. 

\subsection{Uses of Uncertainty in Software Engineering}

Uncertainty quantification is also popular in software engineering, especially in the context of cyber-physical systems. Hu et al.~\cite{xu2022uncertainty} used uncertainty quantification to improve the performance of transfer learning for evolving digital twins of industrial elevators.  
Similarly, the PPT method~\cite{xu2023pretrain} proposes uncertainty-aware transfer learning for digital twins. PPT is evaluated on cyber-physical systems and ADS, with positive results in terms of the effectiveness of uncertainty quantification for reducing the Huber loss in both case studies.

Weiss et al.~\cite{Weiss2021FailSafe} report an empirical study of uncertainty quantification methods that are used to implement supervisors for DNNs. The evaluation is done at the model-level, for four classification classification datasets. Results show that the uncertainty monitors were able to increase the accuracy of the DNNs when supervised. Differently, in this paper, we use uncertainty quantification to inform a system-level failure predictor for ADS.

\subsection{Generic OOD Detectors}

Generic detectors of out-of-distribution samples have been proposed, which we describe for completeness. AutoTrainer~\cite{9402077} analyzes the training process of a DNN to automatically repair when metrics such as accuracy used during training degrade. 
In contrast, UQ operates at testing time, not at training time, to recognize uncertain execution conditions of an ADS, whereas AutoTrainer operates at training time.

Zhang et al.~\cite{9700222} propose an algorithm for the automatic detection of OOD inputs based on the notion of relative activation and deactivation states of a DNN. 
The use of this technique raises some challenges, such as which and how many layers should be selected, and how the different layers should be aggregated. 
SelfChecker~\cite{9402003} helps answer these questions, but the evaluation of the DNN prediction is performed for individual samples. 
UQ works with normal feed-forward passes, making them computationally more efficient and easier to integrate into the ADS development process.

\section{Conclusions and Future Work}\label{sec:conclusions}

In this paper, we describe and evaluate white-box failure predictors based on uncertainty quantification methods. We use them to estimate the confidence of a DNN-based ADS in response to unseen execution contexts. 
Our results show that UQ methods can anticipate many potentially safety-critical failures by several seconds, with a low or zero false alarm rate in anomalous conditions, and a fixed negligible expected false alarm rate in nominal conditions, outperforming two existing solutions from the literature. 

Future work includes extending the comparison to other benchmarks, multi-module ADS, simulators, and ADS case studies such as urban driving for which revisions of the existing methods would be necessary, or alternative confidence score synthesis methods. Furthermore, we intend to broaden our scope by enhancing the detection of more subtle forms of driving quality degradation, such as erratic driving behavior. Additionally, we will explore the implementation of self-healing mechanisms within the simulator and extend our evaluation on physical driving testbeds.

\section*{Acknowledgements}
\addcontentsline{toc}{section}{Acknowledgements}
This research was funded by the Bavarian Ministry of Economic Affairs, Regional Development and Energy.

\balance
\bibliographystyle{IEEEtran}
\bibliography{paper}

\begin{thebibliography}{10}
\providecommand{\url}[1]{#1}
\csname url@samestyle\endcsname
\providecommand{\newblock}{\relax}
\providecommand{\bibinfo}[2]{#2}
\providecommand{\BIBentrySTDinterwordspacing}{\spaceskip=0pt\relax}
\providecommand{\BIBentryALTinterwordstretchfactor}{4}
\providecommand{\BIBentryALTinterwordspacing}{\spaceskip=\fontdimen2\font plus
\BIBentryALTinterwordstretchfactor\fontdimen3\font minus
  \fontdimen4\font\relax}
\providecommand{\BIBforeignlanguage}[2]{{%
\expandafter\ifx\csname l@#1\endcsname\relax
\typeout{** WARNING: IEEEtran.bst: No hyphenation pattern has been}%
\typeout{** loaded for the language `#1'. Using the pattern for}%
\typeout{** the default language instead.}%
\else
\language=\csname l@#1\endcsname
\fi
#2}}
\providecommand{\BIBdecl}{\relax}
\BIBdecl

\bibitem{yurtsever2020survey}
E.~Yurtsever, J.~Lambert, A.~Carballo, and K.~Takeda, ``A survey of autonomous
  driving: Common practices and emerging technologies,'' \emph{IEEE access},
  vol.~8, pp. 58\,443--58\,469, 2020.

\bibitem{grigorescu2020survey}
S.~Grigorescu, B.~Trasnea, T.~Cocias, and G.~Macesanu, ``A survey of deep
  learning techniques for autonomous driving,'' \emph{Journal of Field
  Robotics}, vol.~37, no.~3, pp. 362--386, 2020.

\bibitem{waymos-secret-testing}
``{Waymo Secret Testing},''
  \url{https://www.theatlantic.com/technology/archive/2017/08/inside-waymos-secret-testing-and-simulation-facilities/537648/},
  2017.

\bibitem{Cerf:2018:CSC:3181977.3177753}
\BIBentryALTinterwordspacing
V.~G. Cerf, ``A comprehensive self-driving car test,'' \emph{Commun. ACM},
  vol.~61, no.~2, pp. 7--7, Jan. 2018. [Online]. Available:
  \url{http://doi.acm.org/10.1145/3177753}
\BIBentrySTDinterwordspacing

\bibitem{Gambi:2019:ATS:3293882.3330566}
\BIBentryALTinterwordspacing
A.~Gambi, M.~Mueller, and G.~Fraser, ``Automatically testing self-driving cars
  with search-based procedural content generation,'' in \emph{Proceedings of
  the 28th ACM SIGSOFT International Symposium on Software Testing and
  Analysis}, ser. ISSTA 2019.\hskip 1em plus 0.5em minus 0.4em\relax New York,
  NY, USA: ACM, 2019, pp. 318--328. [Online]. Available:
  \url{http://doi.acm.org/10.1145/3293882.3330566}
\BIBentrySTDinterwordspacing

\bibitem{Abdessalem-ICSE18}
R.~{Ben Abdessalem}, S.~{Nejati}, L.~{C. Briand}, and T.~{Stifter}, ``Testing
  vision-based control systems using learnable evolutionary algorithms,'' in
  \emph{2018 IEEE/ACM 40th International Conference on Software Engineering
  (ICSE)}, May 2018, pp. 1016--1026.

\bibitem{Abdessalem-ASE18-1}
\BIBentryALTinterwordspacing
R.~B. Abdessalem, A.~Panichella, S.~Nejati, L.~C. Briand, and T.~Stifter,
  ``Testing autonomous cars for feature interaction failures using
  many-objective search,'' in \emph{Proceedings of the 33rd ACM/IEEE
  International Conference on Automated Software Engineering}, ser. ASE
  2018.\hskip 1em plus 0.5em minus 0.4em\relax New York, NY, USA: ACM, 2018,
  pp. 143--154. [Online]. Available:
  \url{http://doi.acm.org/10.1145/3238147.3238192}
\BIBentrySTDinterwordspacing

\bibitem{Abdessalem-ASE18-2}
R.~{Ben Abdessalem}, S.~{Nejati}, L.~C. {Briand}, and T.~{Stifter}, ``Testing
  advanced driver assistance systems using multi-objective search and neural
  networks,'' in \emph{2016 31st IEEE/ACM International Conference on Automated
  Software Engineering (ASE)}, Sep. 2016, pp. 63--74.

\bibitem{2020-Riccio-FSE}
V.~Riccio and P.~Tonella, ``{Model-Based Exploration of the Frontier of
  Behaviours for Deep Learning System Testing},'' in \emph{Proceedings of ACM
  Joint European Software Engineering Conference and Symposium on the
  Foundations of Software Engineering}, ser. ESEC/FSE '20, 2020.

\bibitem{NIPS2015_86df7dcf}
D.~Sculley, G.~Holt, D.~Golovin, E.~Davydov, T.~Phillips, D.~Ebner,
  V.~Chaudhary, M.~Young, J.-F. Crespo, and D.~Dennison, ``Hidden technical
  debt in machine learning systems,'' in \emph{Advances in Neural Information
  Processing Systems}, C.~Cortes, N.~Lawrence, D.~Lee, M.~Sugiyama, and
  R.~Garnett, Eds., vol.~28.\hskip 1em plus 0.5em minus 0.4em\relax Curran
  Associates, Inc., 2015.

\bibitem{he2023survey}
W.~He and Z.~Jiang, ``A survey on uncertainty quantification methods for deep
  neural networks: An uncertainty source perspective,'' 2023.

\bibitem{Weiss2021FailSafe}
M.~Weiss and P.~Tonella, ``Fail-safe execution of deep learning based systems
  through uncertainty monitoring,'' in \emph{IEEE 14th International Conference
  on Software Testing, Validation and Verification}, ser. ICST '21.\hskip 1em
  plus 0.5em minus 0.4em\relax IEEE, 2021.

\bibitem{Henriksson}
J.~Henriksson, C.~Berger, M.~Borg, L.~Tornberg, C.~Englund, S.~R.
  Sathyamoorthy, and S.~Ursing, ``Towards structured evaluation of deep neural
  network supervisors,'' in \emph{2019 {IEEE} International Conference On
  Artificial Intelligence Testing ({AITest})}.\hskip 1em plus 0.5em minus
  0.4em\relax {IEEE}, Apr. 2019.

\bibitem{Kim:2019:GDL:3339505.3339634}
\BIBentryALTinterwordspacing
J.~Kim, R.~Feldt, and S.~Yoo, ``Guiding deep learning system testing using
  surprise adequacy,'' in \emph{Proceedings of the 41st International
  Conference on Software Engineering}, ser. ICSE '19.\hskip 1em plus 0.5em
  minus 0.4em\relax Piscataway, NJ, USA: IEEE Press, 2019, pp. 1039--1049.
  [Online]. Available: \url{https://doi.org/10.1109/ICSE.2019.00108}
\BIBentrySTDinterwordspacing

\bibitem{9402003}
Y.~Xiao, I.~Beschastnikh, D.~S. Rosenblum, C.~Sun, S.~Elbaum, Y.~Lin, and J.~S.
  Dong, ``Self-checking deep neural networks in deployment,'' in \emph{2021
  IEEE/ACM 43rd International Conference on Software Engineering (ICSE)}, 2021,
  pp. 372--384.

\bibitem{deeproad}
\BIBentryALTinterwordspacing
M.~Zhang, Y.~Zhang, L.~Zhang, C.~Liu, and S.~Khurshid, ``Deeproad: Gan-based
  metamorphic testing and input validation framework for autonomous driving
  systems,'' in \emph{Proceedings of the 33rd ACM/IEEE International Conference
  on Automated Software Engineering}, ser. ASE 2018.\hskip 1em plus 0.5em minus
  0.4em\relax New York, NY, USA: ACM, 2018, pp. 132--142. [Online]. Available:
  \url{http://doi.acm.org/10.1145/3238147.3238187}
\BIBentrySTDinterwordspacing

\bibitem{dissector}
H.~Wang, J.~Xu, C.~Xu, X.~Ma, and J.~Lu, ``Dissector: Input validation for deep
  learning applications by crossing-layer dissection,'' in \emph{2020 IEEE/ACM
  42nd International Conference on Software Engineering (ICSE)}, 2020, pp.
  727--738.

\bibitem{2020-Stocco-ICSE}
A.~Stocco, M.~Weiss, M.~Calzana, and P.~Tonella, ``Misbehaviour prediction for
  autonomous driving systems,'' in \emph{Proceedings of 42nd International
  Conference on Software Engineering}, ser. ICSE '20.\hskip 1em plus 0.5em
  minus 0.4em\relax ACM, 2020, p. 12 pages.

\bibitem{knoll-monitoring}
\BIBentryALTinterwordspacing
F.~Hell, G.~Hinz, F.~Liu, S.~Goyal, K.~Pei, T.~Lytvynenko, A.~Knoll, and
  C.~Yiqiang, ``Monitoring perception reliability in autonomous driving:
  Distributional shift detection for estimating the impact of input data on
  prediction accuracy,'' in \emph{Computer Science in Cars Symposium}, ser.
  CSCS '21.\hskip 1em plus 0.5em minus 0.4em\relax New York, NY, USA:
  Association for Computing Machinery, 2021. [Online]. Available:
  \url{https://doi.org/10.1145/3488904.3493382}
\BIBentrySTDinterwordspacing

\bibitem{DeepGuard}
\BIBentryALTinterwordspacing
M.~Hussain, N.~Ali, and J.-E. Hong, ``Deepguard: A framework for safeguarding
  autonomous driving systems from inconsistent behaviour,'' \emph{Automated
  Software Engg.}, vol.~29, no.~1, may 2022. [Online]. Available:
  \url{https://doi.org/10.1007/s10515-021-00310-0}
\BIBentrySTDinterwordspacing

\bibitem{2020-Riccio-EMSE}
V.~Riccio, G.~Jahangirova, A.~Stocco, N.~Humbatova, M.~Weiss, and P.~Tonella,
  ``{Testing Machine Learning based Systems: A Systematic Mapping},''
  \emph{Empirical Software Engineering}, 2020.

\bibitem{2020-Humbatova-ICSE}
N.~Humbatova, G.~Jahangirova, G.~Bavota, V.~Riccio, A.~Stocco, and P.~Tonella,
  ``Taxonomy of real faults in deep learning systems,'' in \emph{Proceedings of
  42nd International Conference on Software Engineering}, ser. ICSE '20.\hskip
  1em plus 0.5em minus 0.4em\relax ACM, 2020, p. 12 pages.

\bibitem{2022-Stocco-ASE}
A.~Stocco, P.~J. Nunes, M.~d'Amorim, and P.~Tonella, ``{ThirdEye}: Attention
  maps for safe autonomous driving systems,'' in \emph{Proceedings of 37th
  IEEE/ACM International Conference on Automated Software Engineering}, ser.
  ASE '22.\hskip 1em plus 0.5em minus 0.4em\relax IEEE/ACM, 2022.

\bibitem{udacity-simulator}
{Udacity}, ``{A self-driving car simulator built with Unity},''
  \url{https://github.com/udacity/self-driving-car-sim}, 2017, online; accessed
  25 October 2023.

\bibitem{replication-package}
``Replication package.''
  \url{https://github.com/ast-fortiss-tum/misbehaviour-prediction-with-uncertainty-quantification},
  2023.

\bibitem{nvidia-dave2}
\BIBentryALTinterwordspacing
M.~Bojarski, D.~D. Testa, D.~Dworakowski, B.~Firner, B.~Flepp, P.~Goyal, L.~D.
  Jackel, M.~Monfort, U.~Muller, J.~Zhang, X.~Zhang, J.~Zhao, and K.~Zieba,
  ``End to end learning for self-driving cars.'' \emph{CoRR}, vol.
  abs/1604.07316, 2016. [Online]. Available:
  \url{http://arxiv.org/abs/1604.07316}
\BIBentrySTDinterwordspacing

\bibitem{precrashreport}
N.~H. T. S.~A. U.S. Department~of Transportation, ``Pre-crash scenario typology
  for crash avoidance research,'' 2007.

\bibitem{sotif}
T.~R. I.~. International Organization~for Standardization, ``Road vehicles -
  safety of the intended functionality,'' 2019.

\bibitem{Michelmore}
R.~Michelmore, M.~Kwiatkowska, and Y.~Gal, ``Evaluating uncertainty
  quantification in end-to-end autonomous driving control,'' \emph{CoRR}, vol.
  abs/1811.06817, 2018.

\bibitem{MichelmoreWLCGK20}
\BIBentryALTinterwordspacing
R.~Michelmore, M.~Wicker, L.~Laurenti, L.~Cardelli, Y.~Gal, and M.~Kwiatkowska,
  ``Uncertainty quantification with statistical guarantees in end-to-end
  autonomous driving control,'' in \emph{2020 {IEEE} International Conference
  on Robotics and Automation, {ICRA} 2020, Paris, France, May 31 - August 31,
  2020}.\hskip 1em plus 0.5em minus 0.4em\relax {IEEE}, 2020, pp. 7344--7350.
  [Online]. Available: \url{https://doi.org/10.1109/ICRA40945.2020.9196844}
\BIBentrySTDinterwordspacing

\bibitem{hullermeier2021aleatoric}
E.~H{\"u}llermeier and W.~Waegeman, ``Aleatoric and epistemic uncertainty in
  machine learning: An introduction to concepts and methods,'' \emph{Machine
  Learning}, vol. 110, pp. 457--506, 2021.

\bibitem{bjarnadottir2019climate}
S.~Bjarnadottir, Y.~Li, and M.~G. Stewart, ``Climate adaptation for housing in
  hurricane regions,'' in \emph{Climate Adaptation Engineering}.\hskip 1em plus
  0.5em minus 0.4em\relax Elsevier, 2019, pp. 271--299.

\bibitem{Weiss2021UncertaintyWizard}
M.~Weiss and P.~Tonella, ``Uncertainty-wizard: Fast and user-friendly neural
  network uncertainty quantification,'' in \emph{2021 14th IEEE Conference on
  Software Testing, Verification and Validation (ICST)}.\hskip 1em plus 0.5em
  minus 0.4em\relax IEEE, 2021, pp. 436--441.

\bibitem{10.5555/3045390.3045502}
Y.~Gal and Z.~Ghahramani, ``Dropout as a bayesian approximation: Representing
  model uncertainty in deep learning,'' in \emph{Proceedings of the 33rd
  International Conference on International Conference on Machine Learning -
  Volume 48}, ser. ICML '16.\hskip 1em plus 0.5em minus 0.4em\relax JMLR.org,
  2016.

\bibitem{abdar2021review}
M.~Abdar, F.~Pourpanah, S.~Hussain, D.~Rezazadegan, L.~Liu, M.~Ghavamzadeh,
  P.~Fieguth, X.~Cao, A.~Khosravi, U.~R. Acharya \emph{et~al.}, ``A review of
  uncertainty quantification in deep learning: Techniques, applications and
  challenges,'' \emph{Information fusion}, vol.~76, pp. 243--297, 2021.

\bibitem{2021-Stocco-JSEP}
\BIBentryALTinterwordspacing
A.~Stocco and P.~Tonella, ``Confidence-driven weighted retraining for
  predicting safety-critical failures in autonomous driving systems,''
  \emph{Journal of Software: Evolution and Process}, 2021. [Online]. Available:
  \url{https://doi.org/10.1002/smr.2386}
\BIBentrySTDinterwordspacing

\bibitem{lakshminarayanan2017simple}
B.~Lakshminarayanan, A.~Pritzel, and C.~Blundell, ``Simple and scalable
  predictive uncertainty estimation using deep ensembles,'' \emph{Advances in
  neural information processing systems}, vol.~30, 2017.

\bibitem{fort2019deep}
S.~Fort, H.~Hu, and B.~Lakshminarayanan, ``Deep ensembles: A loss landscape
  perspective,'' \emph{arXiv preprint arXiv:1912.02757}, 2019.

\bibitem{deepxplore}
\BIBentryALTinterwordspacing
K.~Pei, Y.~Cao, J.~Yang, and S.~Jana, ``Deepxplore: Automated whitebox testing
  of deep learning systems,'' in \emph{Proceedings of the 26th Symposium on
  Operating Systems Principles}, ser. SOSP '17.\hskip 1em plus 0.5em minus
  0.4em\relax New York, NY, USA: ACM, 2017, pp. 1--18. [Online]. Available:
  \url{http://doi.acm.org/10.1145/3132747.3132785}
\BIBentrySTDinterwordspacing

\bibitem{deeptest}
\BIBentryALTinterwordspacing
Y.~Tian, K.~Pei, S.~Jana, and B.~Ray, ``Deeptest: Automated testing of
  deep-neural-network-driven autonomous cars,'' in \emph{Proceedings of the
  40th International Conference on Software Engineering}, ser. ICSE '18.\hskip
  1em plus 0.5em minus 0.4em\relax New York, NY, USA: ACM, 2018, pp. 303--314.
  [Online]. Available: \url{http://doi.acm.org/10.1145/3180155.3180220}
\BIBentrySTDinterwordspacing

\bibitem{2021-Jahangirova-ICST}
G.~Jahangirova, A.~Stocco, and P.~Tonella, ``Quality metrics and oracles for
  autonomous vehicles testing,'' in \emph{Proceedings of 14th IEEE
  International Conference on Software Testing, Verification and Validation},
  ser. ICST '21.\hskip 1em plus 0.5em minus 0.4em\relax IEEE, 2021.

\bibitem{deepcrime}
\BIBentryALTinterwordspacing
N.~Humbatova, G.~Jahangirova, and P.~Tonella, ``Deepcrime: Mutation testing of
  deep learning systems based on real faults,'' in \emph{Proceedings of the
  30th ACM SIGSOFT International Symposium on Software Testing and Analysis},
  ser. ISSTA 2021.\hskip 1em plus 0.5em minus 0.4em\relax New York, NY, USA:
  Association for Computing Machinery, 2021, p. 67–78. [Online]. Available:
  \url{https://doi.org/10.1145/3460319.3464825}
\BIBentrySTDinterwordspacing

\bibitem{2024-Lambertenghi-ICST}
S.~C. Lambertenghi and A.~Stocco, ``Assessing quality metrics for neural
  reality gap input mitigation in autonomous driving testing,'' in
  \emph{Proceedings of 17th IEEE International Conference on Software Testing,
  Verification and Validation}, ser. ICST '24.\hskip 1em plus 0.5em minus
  0.4em\relax IEEE, 2024, p. 12 pages.

\bibitem{2020-Haq-ICST}
F.~U. Haq, D.~Shin, S.~Nejati, and L.~Briand, ``Comparing offline and online
  testing of deep neural networks: An autonomous car case study,'' in
  \emph{Proceedings of 13th IEEE International Conference on Software Testing,
  Verification and Validation}, ser. ICST '20.\hskip 1em plus 0.5em minus
  0.4em\relax IEEE, 2020.

\bibitem{12233}
\BIBentryALTinterwordspacing
G.~Lou, Y.~Deng, X.~Zheng, M.~Zhang, and T.~Zhang, ``Investigation into the
  state-of-the-practice autonomous driving testing,'' 2021. [Online].
  Available: \url{https://arxiv.org/abs/2106.12233}
\BIBentrySTDinterwordspacing

\bibitem{2022-Stocco-TSE}
A.~Stocco, B.~Pulfer, and P.~Tonella, ``{Mind the Gap! A Study on the
  Transferability of Virtual vs Physical-world Testing of Autonomous Driving
  Systems},'' \emph{IEEE Transactions on Software Engineering}, 2022.

\bibitem{2023-Stocco-EMSE}
\BIBentryALTinterwordspacing
------, ``Model vs system level testing of autonomous driving systems: A
  replication and extension study,'' \emph{Empirical Softw. Engg.}, vol.~28,
  no.~3, may 2023. [Online]. Available:
  \url{https://doi.org/10.1007/s10664-023-10306-x}
\BIBentrySTDinterwordspacing

\bibitem{10-million-miles}
{BGR Media, LLC}, ``{Waymo's self-driving cars hit 10 million miles},''
  \url{https://techcrunch.com/2018/10/10/waymos-self-driving-cars-hit-10-million-miles},
  2018, online; accessed 25 October 2023.

\bibitem{waymo-driver}
``{Waymo Driver},'' \url{https://waymo.com/waymo-driver/}, 2021.

\bibitem{unity}
``Unity3d.'' \url{https://unity.com}, 2021.

\bibitem{2020-Stocco-GAUSS}
A.~Stocco and P.~Tonella, ``Towards anomaly detectors that learn
  continuously,'' in \emph{Proceedings of 31st International Symposium on
  Software Reliability Engineering Workshops}, ser. ISSREW 2020.\hskip 1em plus
  0.5em minus 0.4em\relax IEEE, 2020.

\bibitem{An2015VariationalAB}
J.~An and S.~Cho, ``Variational autoencoder based anomaly detection using
  reconstruction probability,'' 2015.

\bibitem{Kingma2014AdamAM}
\BIBentryALTinterwordspacing
D.~P. Kingma and J.~Ba, ``Adam: A method for stochastic optimization,''
  \emph{CoRR}, vol. abs/1412.6980, 2014. [Online]. Available:
  \url{https://api.semanticscholar.org/CorpusID:6628106}
\BIBentrySTDinterwordspacing

\bibitem{https://doi.org/10.1002/asi.4630300621}
\BIBentryALTinterwordspacing
D.~C. Blair, ``Information retrieval, 2nd ed. c.j. van rijsbergen. london:
  Butterworths; 1979: 208 pp. price: \$32.50,'' \emph{Journal of the American
  Society for Information Science}, vol.~30, no.~6, pp. 374--375, 1979.
  [Online]. Available:
  \url{https://asistdl.onlinelibrary.wiley.com/doi/abs/10.1002/asi .4630300621}
\BIBentrySTDinterwordspacing

\bibitem{mprofile}
\BIBentryALTinterwordspacing
T.~Palpant, ``{mprofile},'' 1 2023. [Online]. Available:
  \url{https://pypi.org/project/mprofile/}
\BIBentrySTDinterwordspacing

\bibitem{Wilcoxon1945}
\BIBentryALTinterwordspacing
F.~Wilcoxon, ``Individual comparisons by ranking methods,'' \emph{Biometrics
  Bulletin}, vol.~1, no.~6, p.~80, Dec. 1945. [Online]. Available:
  \url{https://doi.org/10.2307/3001968}
\BIBentrySTDinterwordspacing

\bibitem{cohen1988statistical}
J.~Cohen, \emph{Statistical power analysis for the behavioral sciences}.\hskip
  1em plus 0.5em minus 0.4em\relax Hillsdale, N.J: L. Erlbaum Associates, 1988.

\bibitem{carla}
\BIBentryALTinterwordspacing
A.~Dosovitskiy, G.~Ros, F.~Codevilla, A.~L{\'{o}}pez, and V.~Koltun, ``{CARLA:}
  an open urban driving simulator,'' \emph{CoRR}, vol. abs/1711.03938, 2017.
  [Online]. Available: \url{http://arxiv.org/abs/1711.03938}
\BIBentrySTDinterwordspacing

\bibitem{rong2020lgsvl}
G.~Rong, B.~H. Shin, H.~Tabatabaee, Q.~Lu, S.~Lemke, M.~Mo{\v{z}}eiko,
  E.~Boise, G.~Uhm, M.~Gerow, S.~Mehta \emph{et~al.}, ``Lgsvl simulator: A high
  fidelity simulator for autonomous driving,'' \emph{arXiv preprint
  arXiv:2005.03778}, 2020.

\bibitem{beamng}
{BeamNG GmbH}, ``{BeamNG.research},'' \url{https://beamng.tech/}, 2018, online;
  accessed 25 October 2023.

\bibitem{biagiola2023better}
M.~Biagiola, A.~Stocco, V.~Riccio, and P.~Tonella, ``Two is better than one:
  Digital siblings to improve autonomous driving testing,'' 2023.

\bibitem{prescan}
S.~D.~I. Software, ``Simcenter prescan,''
  \url{https://www.plm.automation.siemens.com/global/en/products/simcenter/prescan.html},
  2023.

\bibitem{pro-sivic}
E.~Group, ``Esi prosivic,''
  \url{https://myesi.esi-group.com/downloads/software-downloads/pro-sivic-2021.0},
  2021.

\bibitem{vissim}
{VISSIM}, ``{VISSIM website},''
  \url{https://www.ptvgroup.com/en-us/products/ptv-vissim}, 2023.

\bibitem{VGGNet}
K.~Simonyan and A.~Zisserman, ``Very deep convolutional networks for
  large-scale image recognition.''

\bibitem{arxiv.2204.07874}
\BIBentryALTinterwordspacing
M.~Borg, J.~Henriksson, K.~Socha, O.~Lennartsson, E.~S. Lönegren, T.~Bui,
  P.~Tomaszewski, S.~R. Sathyamoorthy, S.~Brink, and M.~H. Moghadam, ``Ergo,
  smirk is safe: A safety case for a machine learning component in a pedestrian
  automatic emergency brake system,'' 2022. [Online]. Available:
  \url{https://arxiv.org/abs/2204.07874}
\BIBentrySTDinterwordspacing

\bibitem{Strickland-ICRA-2018}
M.~Strickland, G.~Fainekos, and H.~{Ben Amor}, ``\BIBforeignlanguage{English
  (US)}{Deep predictive models for collision risk assessment in autonomous
  driving},'' in \emph{\BIBforeignlanguage{English (US)}{2018 IEEE
  International Conference on Robotics and Automation, ICRA 2018}}, ser.
  Proceedings - IEEE International Conference on Robotics and Automation.\hskip
  1em plus 0.5em minus 0.4em\relax Institute of Electrical and Electronics
  Engineers Inc., 9 2018, pp. 4685--4692.

\bibitem{ayerdi2023metamorphic}
J.~Ayerdi, A.~Iriarte, P.~Valle, I.~Roman, M.~Illarramendi, and A.~Arrieta,
  ``Metamorphic runtime monitoring of autonomous driving systems,'' 2023.

\bibitem{Bogdoll}
\BIBentryALTinterwordspacing
D.~Bogdoll, M.~Nitsche, and J.~M. Zöllner, ``Anomaly detection in autonomous
  driving: A survey,'' 2022. [Online]. Available:
  \url{https://arxiv.org/abs/2204.07974}
\BIBentrySTDinterwordspacing

\bibitem{xu2022uncertainty}
Q.~Xu, S.~Ali, T.~Yue, and M.~Arratibel, ``Uncertainty-aware transfer learning
  to evolve digital twins for industrial elevators,'' in \emph{Proceedings of
  the 30th ACM Joint European Software Engineering Conference and Symposium on
  the Foundations of Software Engineering}, 2022, pp. 1257--1268.

\bibitem{xu2023pretrain}
Q.~Xu, T.~Yue, S.~Ali, and M.~Arratibel, ``Pretrain, prompt, and transfer:
  Evolving digital twins for time-to-event analysis in cyber-physical
  systems,'' \emph{arXiv preprint arXiv:2310.00032}, 2023.

\bibitem{9402077}
X.~Zhang, J.~Zhai, S.~Ma, and C.~Shen, ``Autotrainer: An automatic dnn training
  problem detection and repair system,'' in \emph{2021 IEEE/ACM 43rd
  International Conference on Software Engineering (ICSE)}, 2021, pp. 359--371.

\bibitem{9700222}
Z.~Zhang, P.~Wu, Y.~Chen, and J.~Su, ``Out-of-distribution detection through
  relative activation-deactivation abstractions,'' in \emph{2021 IEEE 32nd
  International Symposium on Software Reliability Engineering (ISSRE)}, 2021,
  pp. 150--161.

\end{thebibliography}

\end{document}